\documentclass{article}

\usepackage{microtype}
\usepackage{graphicx}
\usepackage{subfigure}
\usepackage{booktabs} 
\usepackage[toc,page,header]{appendix}
\usepackage{hyperref}

\def\method{\textsc{ExLM}}

\def\mt{\texttt{[MASK]}}

\usepackage{multirow}
\usepackage[table]{xcolor}
\definecolor{mygray}{gray}{.9}
\definecolor{myred}{RGB}{255,0,0}
\usepackage[figuresright]{rotating}

\usepackage[accepted]{icml2025}

\usepackage{amsmath}
\usepackage{amssymb}
\usepackage{mathtools}
\usepackage{amsthm}

\usepackage[capitalize,noabbrev]{cleveref}

\theoremstyle{plain}

\theoremstyle{definition}

\theoremstyle{remark}

\usepackage{minitoc}

\usepackage[textsize=tiny]{todonotes}

\icmltitlerunning{\method: Rethinking the Impact of \mt{} Tokens in Masked Language Models}

\begin{document}

\doparttoc 
\faketableofcontents 

\twocolumn[
\icmltitle{\method: Rethinking the Impact of \mt{} Tokens in Masked Language Models}



\icmlsetsymbol{equal}{*}
\icmlsetsymbol{corresponding}{\#}

\begin{icmlauthorlist}
\icmlauthor{Kangjie Zheng}{pku}
\icmlauthor{Junwei Yang}{pku}
\icmlauthor{Siyue Liang}{pku}
\icmlauthor{Bin Feng}{idea,pku}
\icmlauthor{Zequn Liu}{pku}

\icmlauthor{Wei Ju}{scu} 
\icmlauthor{Zhiping Xiao}{corresponding,uw} 
\icmlauthor{Ming Zhang}{corresponding,pku}

\end{icmlauthorlist}


\icmlaffiliation{pku}{State Key Laboratory for Multimedia Information Processing, School of Computer Science, PKU-Anker LLM Lab, Peking University, China.
}

\icmlaffiliation{idea}{International Digital Economy Academy (IDEA), Shenzhen, China.}

\icmlaffiliation{scu}{College of Computer Science, Sichuan University, Chengdu, China.}
\icmlaffiliation{uw}{Paul G. Allen School of Computer Science and Engineering, University of Washington, U.S.A}

\icmlcorrespondingauthor{Ming Zhang}{mzhang\_cs@pku.edu.cn}

\icmlcorrespondingauthor{Zhiping Xiao}{patxiao@uw.edu}
\icmlkeywords{Masked Language Model, Language Modeling}

\vskip 0.3in
]

\printAffiliationsAndNotice{}

\begin{abstract}

Masked Language Models (MLMs) have achieved remarkable success in many self-supervised representation learning tasks. MLMs are trained by randomly masking portions of the input sequences with \mt{} tokens and learning to reconstruct the original content based on the remaining context. This paper explores the impact of \mt{} tokens on MLMs. Analytical studies show that masking tokens can introduce the \textbf{\textit{corrupted semantics}} problem, wherein the corrupted context may convey multiple, ambiguous meanings. This problem is also a key factor affecting the performance of MLMs on downstream tasks. Based on these findings, we propose a novel enhanced-context MLM, \method{}. Our approach expands \mt{} tokens in the input context and models the dependencies between these expanded states. This enhancement increases context capacity and enables the model to capture richer semantic information, effectively mitigating the corrupted semantics problem during pre-training. Experimental results demonstrate that \method{} achieves significant performance improvements in both text modeling and SMILES modeling tasks. Further analysis confirms that \method{} enriches semantic representations through context enhancement, and effectively reduces the semantic multimodality commonly observed in MLMs. 

\end{abstract}

\begin{figure}[t]
\centering
	\includegraphics[width=0.95\linewidth]{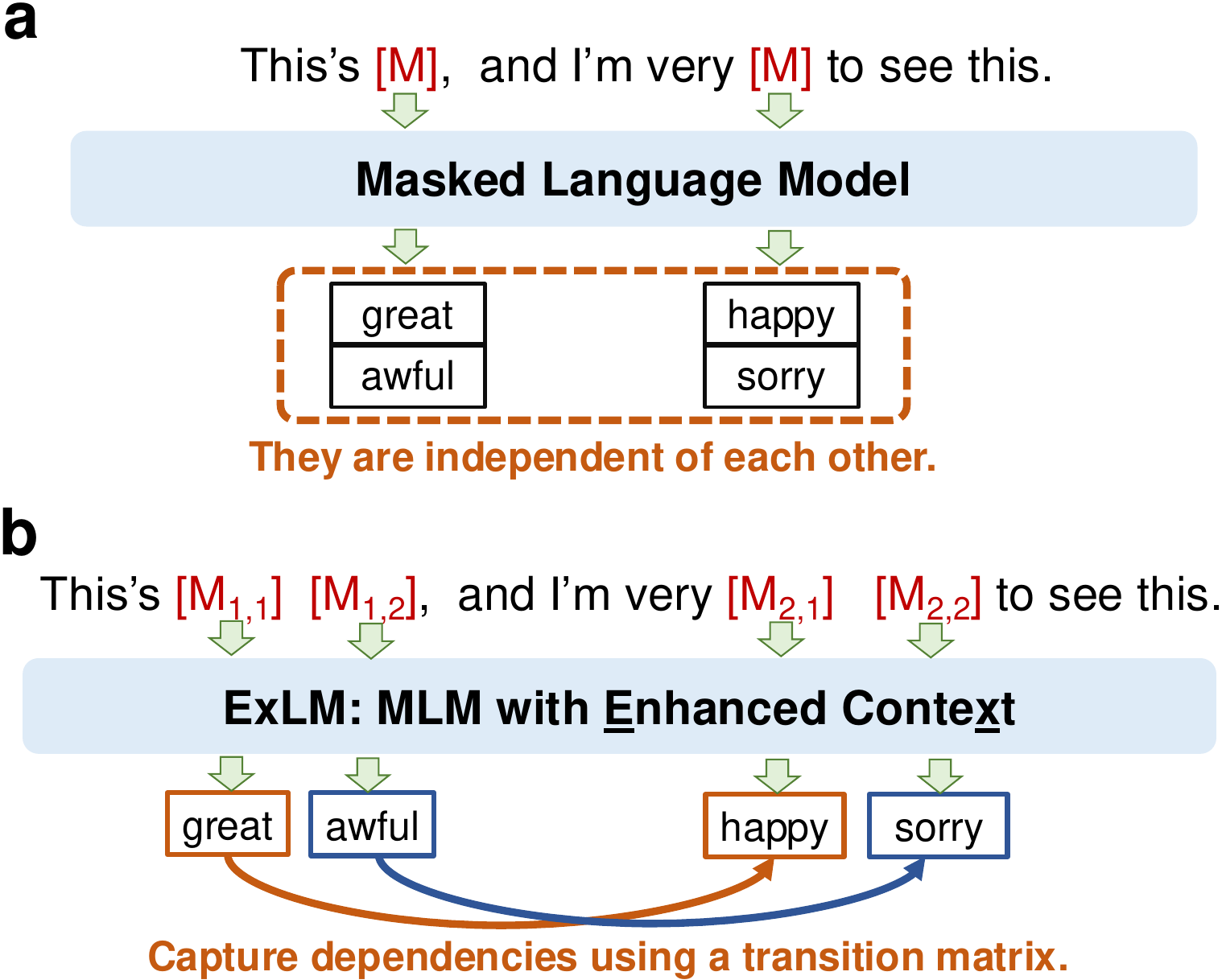}
	\caption {Illustrations of the vanilla MLM (\textbf{a}) and \method{} (\textbf{b}). MLM can be affected by the multimodality problem. In \method{}, the model creates multiple hidden states for each \mt{} token (e.g., $[\texttt{M}_{1,1}], [\texttt{M}_{1,2}], [\texttt{M}_{2,1}], [\texttt{M}_{2,2}]$). By leveraging a larger semantic space and explicitly modeling the dependencies between these states, the model can capture richer semantic information in the enhanced context while mitigating the effects of multimodality.}
\label{fig::exmlm_vsmlm}
\end{figure}

\section{Introduction}
\label{introduction}

Pre-trained masked language models (MLMs) have achieved significant success across various types of sequence data, including text \citep{devlin2018bert, liu2019roberta, lan2019albert, he2020deberta, joshi2020spanbert, meng2023representation}, small molecules \citep{wang2019smiles, ross2022large, pan2023large, zheng2024smi}, and proteins \citep{ lin2022language, lin2023evolutionary, su2023saprot, zhengesm, hayes2025simulating}, establishing themselves as a foundational approach for sequence representation learning tasks. To enable the effective extraction of useful semantic information, MLM employs a mask-then-predict training strategy. During the pre-training process, a certain proportion (typically $15\%$) of tokens in the input sequence are randomly replaced with a special symbol, \mt{}. The model is then trained to predict the original tokens based on the corrupted context. This process allows MLMs to learn meaningful semantic information in a self-supervised manner from large-scale unlabeled data. The learned knowledge can be effectively transferred to a wide range of downstream tasks, such as text classification and molecular property prediction, significantly improving the performance of deep learning models in these domains.

Although MLMs have achieved significant success in numerous tasks and applications, their effectiveness remains an important research problem. During the MLM training process, parts of the input are masked, and unreal tokens (\mt{}) are introduced into the context. This process impacts the input context of MLMs in two critical aspects:

\begin{itemize}
    \item Introducing \textbf{\textit{Unreal Tokens}}: The context provided to MLM during pre-training contains a large number of unreal tokens (\mt{}) that are absent from real-world text, potentially distorting the learning process.

    \item Resulting in \textit{\textbf{Corrupted Semantics}}: The replacement of tokens with \mt{} results in incomplete context semantics, which can negatively affect the model's ability to learn accurate semantic representations.
\end{itemize}

These two aspects of impact are closely tied to the mask ratio used during pre-training. Higher mask ratios exacerbate the problems of unreal tokens and corrupted semantics at the same time, leading to a noticeable performance drop when the mask ratio is too high. While previous studies have investigated the unreal token problem and its impact on MLM performance \citep{clark2020electra, meng2023representation}, they have largely overlooked the corrupted semantics problem. Consequently, there is little work systematically exploring the impact of both problems on MLM performance, or evaluating the magnitude of their effect independently. This gap arises because these two factors are interdependent and both rely on the mask ratio, making it challenging to design experiments that disentangle their respective effects. 

\begin{figure}[t]
\centering
	\includegraphics[width=0.95\linewidth]{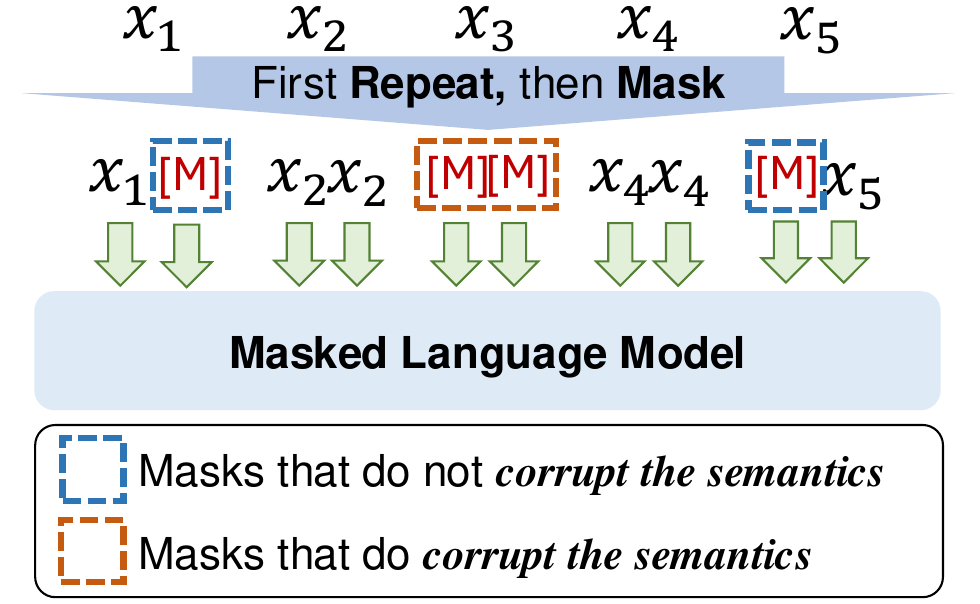}
	\caption {Illustrations of the \textit{Repeated MLM} experiment. Each token in the model input is repeated before being masked. The artificial redundancy introduced into the input ensures that replacing a token with {\mt} ($[\texttt{M}]$) does not necessarily lead to semantic corruption in the context, as the repeated tokens provide additional information to preserve the original semantics. A token is regarded as having corrupted semantics only when all its copies are masked.}
\label{fig::repeat_mlm}
\end{figure}

To fill this research gap, we designed and conducted the \textit{Repeated MLM} analytical experiments (shown in Figure \ref{fig::repeat_mlm}) to separately evaluate the relative impact of these two factors. The experimental results demonstrate that \textbf{the corrupted semantics problem has a significantly greater impact on MLM performance than the unreal tokens problem}. This is further reflected in a stronger semantic multimodality \cite{gu2017non}, where multiple plausible predictions exist for the original tokens due to the ambiguous or context-dependent meanings. Based on these findings, we propose a novel pre-trained model, \method{}, which enhances context representation in MLMs. By expanding each \mt{} token in the input context into multiple hidden states, the model is provided with a larger semantic space. This allows it to capture richer semantic information associated with each \mt{} token, thereby reducing the semantic multimodality in token prediction. Furthermore, we introduce a transition matrix between these expanded states, enabling the model to directly capture semantic relationships among different states. To efficiently train the \method{}, we propose a state alignment algorithm based on dynamic programming, which aligns target tokens with expanded states in a data-driven manner, significantly improving model performance.

In summary, the contributions of this work are as follows:

\begin{itemize}
    \item We conduct the first systematic analysis of MLM behavior from the dual perspectives of both \textit{\textbf{unreal tokens}} and \textit{\textbf{corrupted semantics}}. Through a series of carefully designed experiments, we reveal that the corrupted semantics problem has a greater impact on MLM performance, providing new insights into MLM studies.
    \item Based on the analysis, we propose \method{}, a novel pre-trained model that enhances context representation for MLMs. By expanding \mt{} tokens into multiple states and employing a state alignment algorithm, \method{} effectively improves semantic modeling and reducing the semantic multimodality in the context.
    \item Extensive experimental results on text and molecular property prediction tasks demonstrate the superior performance of \method{}. Further analysis confirms its effectiveness and highlights its potential for addressing challenges in MLM pre-training.
\end{itemize}

\section{Background of MLM}
\label{sec::background}

The pre-training objective of MLMs is to predict missing tokens from a partially masked input sequence. In this process, a portion of the input tokens is randomly selected and masked (i.e., replaced with \mt{}), forming a corrupted sequence. The model's task is to recover the original tokens using the remaining unmasked context as input.

Specifically, let's define a sequence of tokens as \( \mathbf{X} = [x_1, x_2, \dots, x_n] \). In the MLM pre-training process, some of the tokens in \( \mathbf{X} \) are randomly replaced with a special token \mt{}, producing a partially masked input sequence \(  \tilde{\mathbf{X}} = [\tilde{x}_1, \tilde{x}_2, \dots, \tilde{x}_n] \), where:
\[
\tilde{x}_i =
\begin{cases}
[{\rm MASK}] & \text{if token } x_i \text{ is selected to be masked}\,, \\
x_i & \text{otherwise}\,.
\end{cases}
\]
The model is trained to predict the original tokens \( x_i \) corresponding to the masked positions \( i \) in \( \tilde{\mathbf{X}} \). Typically, a fixed percentage of the input tokens (e.g., \( 15\% \) ) are randomly selected to be masked during the pre-training process. The objective of MLM pre-training is to minimize the discrepancy between the predicted tokens and the true tokens at the masked positions. This is achieved by maximizing the likelihood of the true tokens given the context of the unmasked tokens. The objective function for MLM pre-training can be formulated as:
\[
\mathcal{L}_{\text{MLM}} = - \sum_{i \in M} \log P(x_i \mid \tilde{x}_1, \dots, \tilde{x}_n)\,,
\]
where \( M \) denotes the set of indices corresponding to the masked tokens, and \( P(x_i \mid \tilde{x}_1, \dots, \tilde{x}_n) \) represents the model's predicted probability for a certain token \( x_i \) at the masked position \( i \), conditioned on a corrupted context formed by the unmasked tokens in \( \tilde{\mathbf{X}} \).

\begin{figure}[t]
\centering
	\includegraphics[width=0.99\linewidth]{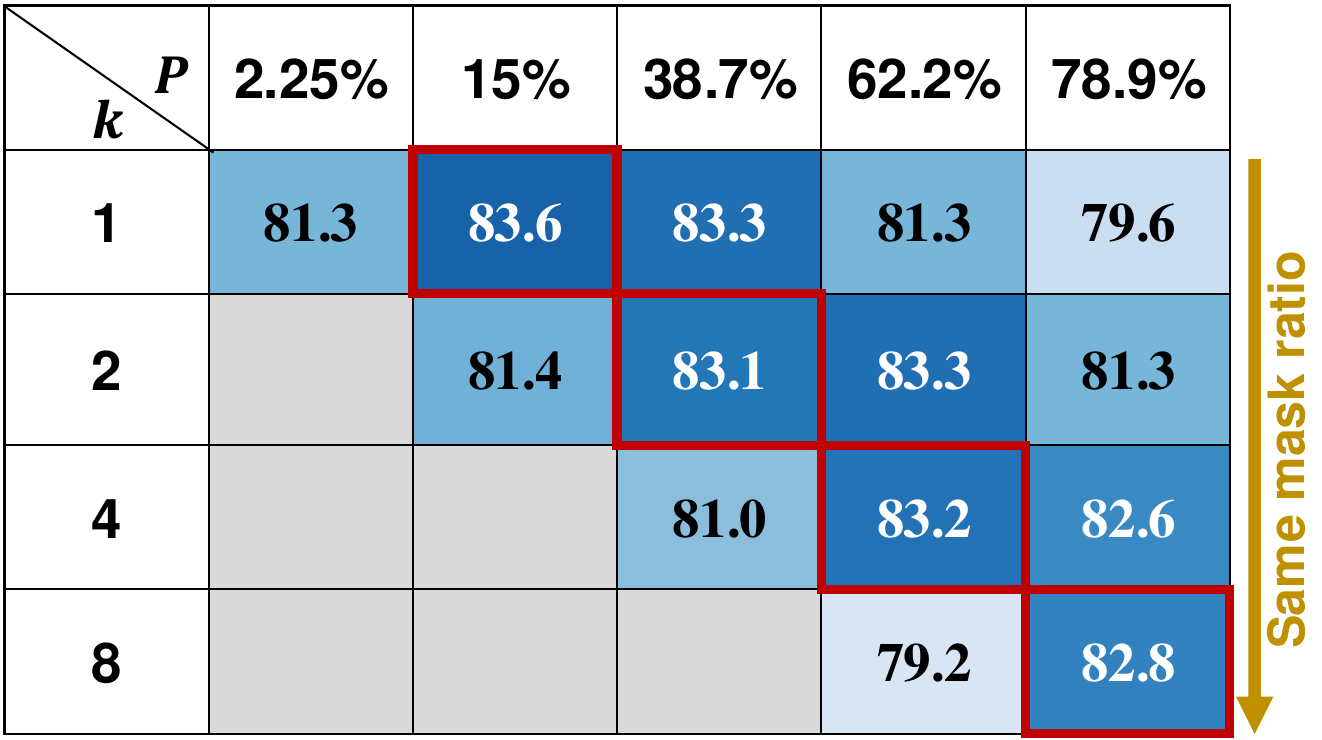}
	\caption {Results of the \textit{Repeated MLM} experiment. These are the evaluation results (i.e., accuracy) of MLMs with different repetition times \(k\) and mask ratios \(p\) on the MNLI task \citep{MNLI}, with results of similar performance highlighted in similar colors.}
\label{fig::analyse_result1}
\end{figure}

\section{Understanding the Impact of \mt{}}
\label{sec::mask_impact}
In this section, we design analytical experiments to explore the impact of \textit{\textbf{unreal tokens}} and \textit{\textbf{corrupted semantics}} on MLM performance. However, since these two factors are strongly interrelated in MLMs, we first designed a \textit{Repeated MLM} experiment to decouple unreal tokens and corrupted semantics (Section \ref{sec::decoupling}). This allows us to separately explore the impact of each factor and demonstrate that corrupted semantics has a significantly greater effect on MLM performance (Section \ref{sec::compare_results_analyse}). Additionally, our analysis further shows that more severe corrupted semantics lead to the loss of critical context semantics, which exacerbates the multimodality in token predictions (Section \ref{sec::uncertain}).

\subsection{Decoupling \textit{Corrupted Semantics} and \textit{Unreal Tokens}}
\label{sec::decoupling}

\begin{figure}[t]
\centering
	\includegraphics[width=0.86\linewidth]{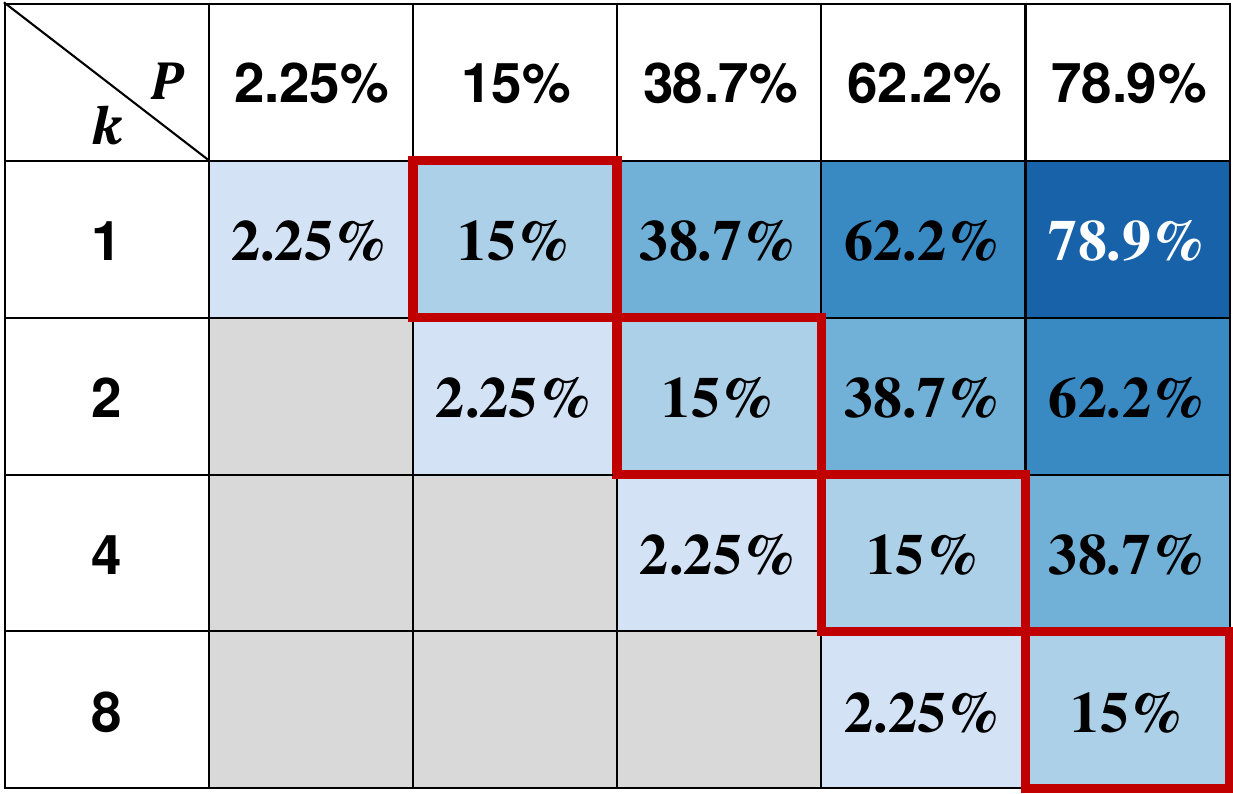}
	\caption {The corrupted semantics proportions corresponding to each set of \textit{Repeated MLM} experiments are reported. When the repetition times are \(k\) and the mask ratio is \(p\), the proportion of corrupted semantics is \(p^k\) (see Appendix \ref{ev_csp} for a detailed proof).}
\label{fig::analyse_csrate}
\end{figure}

In traditional MLM training, the mask ratio plays a crucial role in determining both the proportion of \mt{} tokens introduced into the context and the proportion of semantic information discarded from the original context. This interplay makes it challenging to isolate and compare the impact of these two factors--the proportion of \mt{} tokens 
 (i.e., \textbf{\textit{unreal tokens}}) and the proportion of \textbf{\textit{corrupted semantics}}--on MLM performance. To address this challenge, we designed the \textit{Repeated MLM} experiment.

The \textbf{core idea} of this experiment is to artificially introduce redundancy into the MLM's input, ensuring that replacing a token with \mt{} does not inevitably lead to semantic corruption in the context.

Specifically, as illustrated in Figure \ref{fig::repeat_mlm}, before feeding the sequence into the MLM, we first repeat each token \(k\) times, where \(k \in \mathbb{N}\) is a hyperparameter set based on experimental requirements. We then randomly mask the repeated sequence at a certain ratio \(p\), which means replacing \(p \in (0.0\%, 100.0\%) \) of the tokens with \mt{}. This setup creates an interesting phenomenon: while the proportion of unreal \mt{} tokens in the context remains \(p\) due to the masking ratio, the proportion of corrupted semantics changes. Since each token has \(k\) copies and the probability of each copy being masked is \(p\), the probability of the semantic information carried by a token being completely corrupted becomes \(p^k\). We also have provided a detailed proof about the results in Appendix \ref{ev_csp}.

By keeping \(p\) fixed and varying \(k\), we ensure that the proportion of \mt{} tokens in the context remains constant while adjusting the degree of corrupted semantics. This enables us to control variables effectively, and systematically compare the impact of these two factors on MLM performance by measuring each factor separately.

\subsection{What Matters More: \textit{Corrupted Semantics} or \textit{Unreal Tokens}}
\label{sec::compare_results_analyse}

Under the experimental setup designed in Section \ref{sec::decoupling}, we have trained a series of MLMs with different repetition times \(k\) and mask ratios \(p\) to analyze MLM's behavior. For consistency, all training hyperparameters of the MLMs in the experiments are kept the same except for \(p\) and \(k\). Additionally, during downstream fine-tuning, the input is repeated with the same repetition times \(k\) as in pre-training. More detailed training configurations and hyperparameters can be found in Appendix \ref{sec::analyse_configuration}. The results of the MNLI task, evaluated using accuracy as the primary metric \citep{MNLI}, are presented in Figure \ref{fig::analyse_result1}. We also provide the results of this experiments on more tasks in Appendix \ref{sec::more_repeat_mlm}. Some entries are blank (gray areas) due to overly low proportions of corrupted semantics (less than $0.05\%$), causing very low loss during pre-training and unstable model training problem. For comparison, the proportions of corrupted semantics for each experiment are provided in Figure \ref{fig::analyse_csrate}.

From these results, we can observe that both excessively large and excessively small corrupted semantics lead to significant performance degradation in the model. Besides, \textbf{when the proportions of corrupted semantics remain constant and the mask ratio varies, the performance of the MLM changes only slightly}. Although there is a minor decline in performance as the mask ratio increases (red cells in Figure \ref{fig::analyse_result1}, from $83.6$ to $82.8$), the overall performance remains relatively similar. As long as the proportions of corrupted semantics are not excessively high, the model can still maintain relatively good performance even if the context in the MLM pre-training process contains a large number of \mt{} tokens (e.g., \(p=78.9\%\), \(k=8\)). 

In contrast, when the mask ratio remains fixed and the proportions of corrupted semantics increase, the model's performance exhibits more significant changes (from $82.8$ to $79.6$). This demonstrates that \textbf{the corrupted semantics problem has a more pronounced impact on performance compared to the unreal tokens problem}.

\subsection{Core Impact of \textit{Corrupted Semantics}: Multimodality}
\label{sec::uncertain}

\begin{figure}[t]
\centering
	\includegraphics[width=0.99\linewidth]{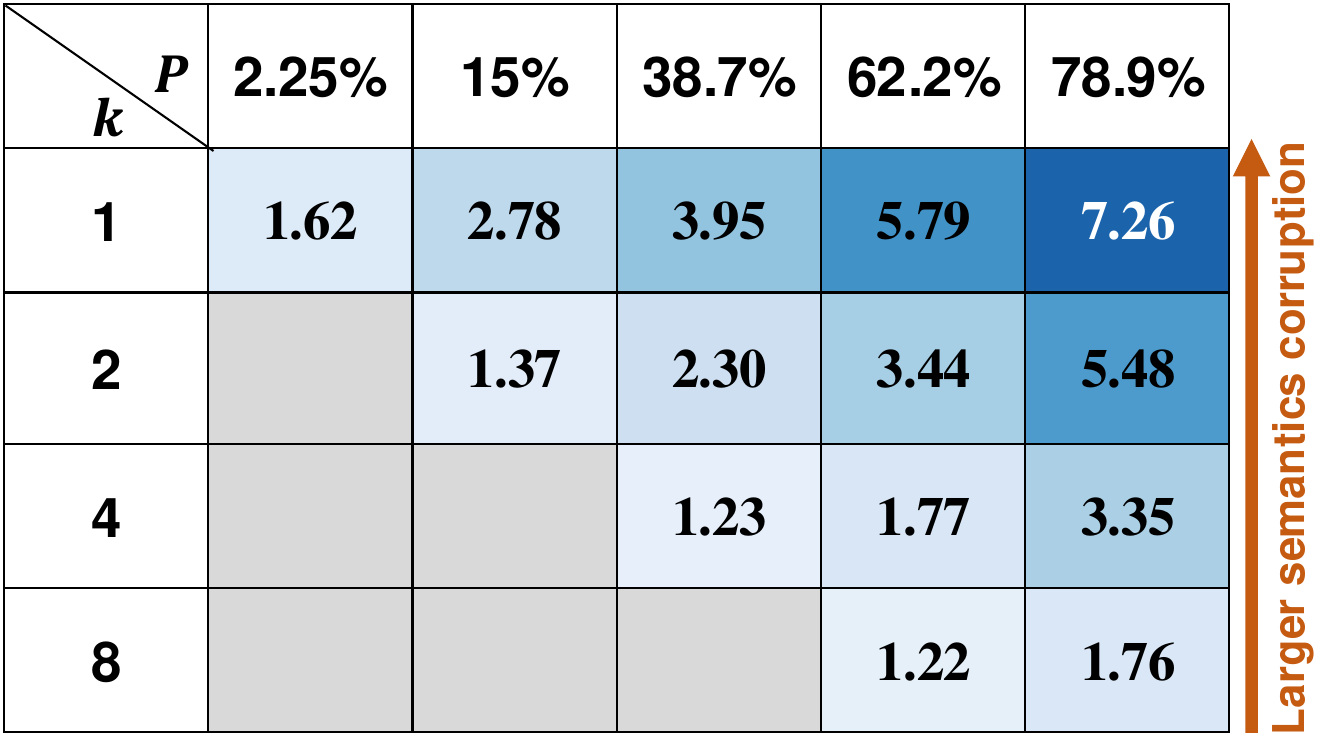}
	\caption {Entropy analysis in the \textit{Repeated MLM} experiment. We visualize the entropy (in bits) of different models during mask prediction. Larger semantics corruption significantly leads to an increase in the entropy of the model's prediction distribution.}
\label{fig::analyse_result_ent}
\end{figure}

We conducted a deeper analysis to investigate how \textbf{\textit{corrupted semantics}} influence the performance of MLMs.
As illustrated in Figure \ref{fig::exmlm_vsmlm}(a), the reconstruction of a \mt{} token relies on the semantic information provided by its context.
When a corrupted context is given, it may imply multiple different potential semantics, resulting in significantly different reconstruction outcomes.
This phenomenon is referred to as multimodality, a concept initially proposed in NAT \citep{gu2017non}.
Multimodality causes MLMs to produce more mixed and uncertain predictions during pre-training, thereby significantly affecting downstream performance.
To gain a more straightforward understanding of multimodality, we analyzed the prediction entropy of MLMs with different repetition times \(k\) and mask ratios \(p\). We also have provided more details on the entropy calculation process in Appendix \ref{app::ent_details}.
As illustrated in Figure \ref{fig::analyse_result_ent}, with an increase in semantic corruption, the prediction entropy also rises, indicating a more severe semantic multimodality. Such multimodal phenomenon also significantly impacts the performance of MLMs.

On the other hand, when corruption is too low (e.g., $2.25\%$), the context still contains plenty of semantic clues, making token prediction too easy. At this point, the model's predictions also exhibit very low uncertainty. The training curve analysis in Appendix \ref{sec::repeat_mlm_curve} also shows that when the corruption is too low, the training task becomes overly simplistic. This simplicity prevents the model from learning deeper knowledge and ultimately reduces performance. 

Therefore, an optimal approach would be to design the input context with a reasonable degree of semantic corruption, ensuring the model can correctly handle the missing semantics and learn meaningful knowledge. This balance would avoid the negative impacts of the semantic multimodality problem while maintaining the model's performance.

\section{Proposed Method: \method}
\label{exmlm_pre_training}

\begin{figure*}[!t]
\centering
\includegraphics[width=1.00\linewidth]{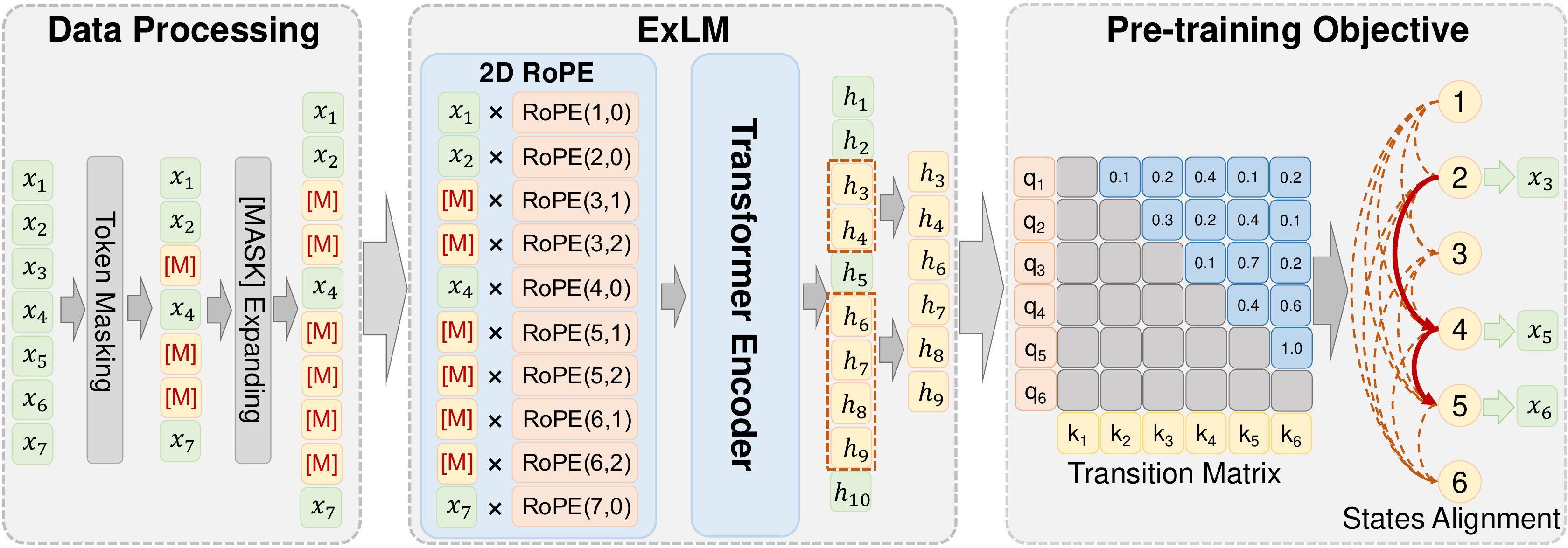}
\caption{Overview of our proposed \method{}. \method{} creates multiple expanded states for each \mt{} token, providing a larger semantic space and a stronger ability to capture the missing semantics in the context. Additionally, it explicitly models the semantic dependencies between the expanded states using a transition matrix, which is computed based on the representation of each state \(\mathbf{h}_i\).}
\label{fig_overview}
\end{figure*}

In this section, we will first introduce the core design concepts and overall architecture of \method{} (Section \ref{overview}). Then, we will elaborate on two key components of context enhancement, \textbf{states expansion} and \textbf{dependency capture}, by detailing the model design and training algorithm for each (Section \ref{expanded_input} and Section \ref{training}).

\subsection{Overview}
\label{overview}

\begingroup
\color{black}


Through the analysis in Section \ref{sec::mask_impact}, we have identified that corrupted semantics is the main factor affecting MLM's performance. When semantics of the context are severely corrupted, the resulting multimodality makes it increasingly difficult for the MLM to restore the original tokens.

Building on this, an intuitive approach to improving the MLM is: \textbf{how can we enhance the model's ability to better handle semantic multimodality?} More specifically, the semantic multimodality can be divided into two aspects:


\begin{itemize}
    \item \textbf{Intra-token Multimodality}: The potential choices for each missing token become more diverse, and the significant semantic differences among these choices increase the semantic diversity and ambiguity.
    \item \textbf{Inter-token Multimodality}: The meaning of one token is intricately linked to and influenced by the meanings of other tokens, resulting in complex semantic interactions and dependencies between missing tokens. 
\end{itemize}

Therefore, we need to enhance the model’s capabilities in two areas: one is improving its ability to model diverse and ambiguous semantics, and the other is enhancing its ability to capture semantic dependencies between missing tokens. Following this core idea, we propose a novel enhanced-context MLM (\method{}) with two main improvements: 

\begin{itemize}
    \item \textbf{States Expansion}: For each \mt{} token in the input context, multiple hidden states are created, enlarging the semantic space for the model. A larger semantic space enables the model to capture richer and more diverse semantic information, better handling the semantic diversity introduced by multimodality.
    \item \textbf{Dependency Capture}: A transition matrix is used to explicitly model the semantic dependencies between different states, and the States Alignment algorithm based on dynamic programming efficiently provides supervision signals to these hidden states.
\end{itemize} 

\endgroup

Therefore, we will first introduce the details of context enhancement, which includes States Expansion and Dependency Capture (Section \ref{expanded_input}), and then present the States Alignment algorithm for \method{} pre-training (Section \ref{training}).

\subsection{Modeling Semantics with Enhanced Context}
\label{expanded_input}

The core of context enhancement process is to create multiple hidden states for each \mt{} token and capture the dependencies among these states, enhancing the model's ability to capture richer semantic information. This process consists of three main steps. First, the model expands each \mt{} token in the input into multiple hidden states, with the number of states determined by a hyperparameter $k$. Secondly, to enable the model to distinguish between these expanded states, it further employs 2D Rotary Position Embedding (RoPE) to differentiate the states. Finally, \method{} uses a transition matrix to explicitly capture the semantic dependencies between these hidden states, enhancing the model's ability to model the missing semantic dependencies.

\paragraph{Expanding \mt{} tokens in the inputs.}

A key idea in \method{} is to duplicate the embeddings of \mt{} tokens, effectively creating multiple “clones” for each \mt{} before feeding them into the model. More formally, for each \mt{} token in the original input sequence $\tilde{\textbf{X}}$ = \(\big[x_1, x_2, \dots, x_i = [{\rm MASK}], \dots, x_n\big]\), we take its embedding \(\mathbf{e}_{[{\rm MASK}]}\) and make \(k\) copies. We then form an expanded input sequence by replacing the single \mt{} token with its \(k\) duplicated embeddings:
\[
\mathbf{X'} = [\mathbf{e}_{x_1}, \mathbf{e}_{x_2}, \dots, \mathbf{e}_{[{\rm MASK}]}^{(1)}, \dots, \mathbf{e}_{[{\rm MASK}]}^{(k)}, \dots, \mathbf{e}_{x_n}],
\]
where $\mathbf{e}_{x_i}$ is the embedding of token $x_i$, $\mathbf{e}_{[{\rm MASK}]}^{(i)}$ is the $i$-th copy of \mt{} embedding \(\mathbf{e}_{[{\rm MASK}]}\), and \(k\) is a hyperparameter controlling the number of cloned embeddings. 

This expanded sequence \(\mathbf{X'}\), consisting of embeddings for both the original tokens and the duplicated \mt{} tokens, is then passed into a Transformer Encoder \(\boldsymbol{\theta}\) for contextual encoding $\mathbf{H}$ \citep{vaswani2017attention}:
\[
\mathbf{H} = [\mathbf{h}_{x_1}, \mathbf{h}_{x_2}, \dots, \mathbf{h}_{[{\rm MASK}]}^{(1)}, \dots, \mathbf{h}_{[{\rm MASK}]}^{(k)}, \dots, \mathbf{h}_{x_n}],
\]
where each \(\mathbf{h}_t\) represents the hidden state corresponding to the input embedding \(\mathbf{e}_t\) in \(\mathbf{X'}\). By explicitly expanding \mt{} tokens, the model is equipped to learn richer and more diverse representations for the missing information, leveraging these enriched embeddings to better capture semantic information and reconstruct the original context.
\paragraph{Using 2D RoPE to distinguish expanded states.}
In the context enhancement process, each \mt{} token is duplicated multiple times, creating several “clones” that the model may find difficult to distinguish. To address this challenge, we introduce a 2D Rotary Position Embedding (RoPE) mechanism \citep{su2021roformer}, which leverages a second dimension in the positional information to differentiate these clones. Specifically, if the original \mt{} token is located at position \(i\) in the sequence, its \(k\) duplicates are assigned unique 2D positions: \((i, 1), (i, 2), \dots, (i, k)\). Meanwhile, all original (non-\mt{}) tokens retain their original positions, represented as \((j, 0)\), where \(j\) denotes their index in the sequence. Here, the first coordinate captures the token’s position in the original sequence, while the second coordinate differentiates the clones of a \mt{} token. Using this two-dimensional positional structure, the 2D RoPE mechanism applies rotational position embeddings to encode both the sequence position and the clone index. 

\paragraph{Using a transition matrix to capture dependencies between expanded states.}
The semantic dependencies between expanded states can be modeled as a directed acyclic graph (DAG), where each node represents a state, and the edge weights indicate the strength of the semantic dependencies between pairs of states. Similar to previous work \citep{huang2022directed}, we adopt a DAG to effectively capture these semantic dependencies.

Specifically, the representation of each state \(\mathbf{h}_i\) extracted by the Transformer Encoder undergoes an attention-like computation to derive the transition matrix \(\mathbf{E}\). This transition matrix \(\mathbf{E}\) serves as the adjacency matrix of the DAG, quantifying the semantic association strength between different states. The computation is defined as follows:
\[
\mathbf{E} = \text{softmax}\left(\frac{\mathbf{Q} \mathbf{K}^\top}{\sqrt{d}} + \mathbf{M}\right), \label{eq:matrixE}
\]
\[
\mathbf{Q} = \mathbf{H} \mathbf{W}_\text{Q}, \quad \mathbf{K} = \mathbf{H} \mathbf{W}_\text{K}, \notag
\]
where \(d\) is the hidden size, \(\mathbf{W}_\text{Q}\) and \(\mathbf{W}_\text{K}\) are learnable weight matrices, \(\mathbf{M}\) is an upper triangular mask matrix ensuring that \(\mathbf{E}\) remains an upper triangular matrix, thereby enforcing the DAG structure by preventing backward edges.

Additionally, each state representation \(\mathbf{h}_i\) is passed through the model's token prediction head to compute the probability distribution over possible tokens for that state:
\[
\mathbf{P} = \text{softmax}(\mathbf{H} \mathbf{W}_\text{P}^\top),
\]
where \(\mathbf{P}\) represents the probability distributions for each state, \(\mathbf{W}_\text{P}\) is a learnable weight matrix for token prediction.

By using the transition matrix \(\mathbf{E}\), the model explicitly captures the semantic dependencies between expanded states, enhancing its ability to reconstruct the missing semantic information effectively.

\subsection{Pre-training of \method{}: States Alignment}
\label{training}

A key challenge in training \method{} lies in the fact that each \mt{} token is expanded into multiple hidden states. This expansion means that there are more hidden states than the target tokens to predict. Consequently, we must determine an alignment between these states and the target tokens—that is, deciding which hidden state should be responsible for predicting which token. This alignment process is at the heart of our States Alignment algorithm.

\begin{table*}[t]
\centering
\caption{The overall results on $7$ molecule property classification datasets. We report ROC-AUC
score (higher is better) under scaffold splitting. The best results are \textbf{bold}. The second-best results are \underline{underlined}. * indicates that the model uses the same training data, model architecture, and training hyperparameters as \method{}. For more detailed information about the dataset, please refer to Table \ref{fine_tune_dataset}. 
}
\footnotesize
\begin{tabular}{c|ccccccc|c}

\toprule
 Datasets    &   BACE$\uparrow$&   BBBP$\uparrow$&   Tox21$\uparrow$&  SIDER$\uparrow$&     MUV$\uparrow$&    ClinTox$\uparrow$&    ToxCast$\uparrow$ & Mean$\uparrow$\\
 \# Molecules &  1531&   2039&   7831&  1427&      93087&     1478&    8575 &-\\
 \midrule
 D-MPNN &   80.9&   71.0&   75.9&  57.0&      \underline{78.6}&       \underline{90.6}&    \underline{65.5} &  \underline{74.2}\\
 Attentive FP &   78.4&   64.3&   76.1&  60.6&     76.6&      84.7&    63.7 & 72.1\\
 N-Gram$_{\mathrm{RF}}$ & 77.9&   69.7&   74.3&  \textbf{66.8}&    76.9&      77.5&  - & - \\
 GROVER&   \textbf{82.6}&   70.0&   74.3 &  \underline{64.8}&  62.5&          81.2&    65.4 & 71.5\\
 GraphMVP &   \underline{81.2}&   \underline{72.4}&   75.9&  63.9&     77.7&       79.1&    63.1 & 73.3\\

Mole-BERT & 80.8&   71.9&   \underline{76.8}&  62.8&     \underline{78.6}&       78.9&    64.3 & 73.4
 \\
 3D InfoMax & 79.7&   69.1&   74.5&  60.6&      74.4&       79.9&    64.4 & 71.8 \\
  
SMILES-BERT$^*$ & 77.8&   68.6&   75.1&  61.2&      75.1&      89.8&    64.9 & 73.2 \\
\midrule 
 \rowcolor{mygray} \method{} & 79.6 & \textbf{72.8} & \textbf{78.2} &   64.5 & \textbf{78.8} & \textbf{91.6} & \textbf{66.9} & \textbf{76.1}\\
\bottomrule
\end{tabular}
\label{table::classification}
\end{table*}

\paragraph{Formulating States Alignment as a DAG decoding problem.} 
Our goal is to maximize the probability of the DAG decoding all target tokens \(\mathbf{Y}\) (i.e., all masked tokens). Formally, let \(\Gamma\) denote the set of all possible paths (i.e., all possible ways to align states to the target tokens). Each path \(\mathbf{A} \in \Gamma\) represents one particular alignment of states to the target tokens in \(\mathbf{Y}\). The training objective can thus be written as a marginalization over all possible alignments:
\[
\mathcal{L}_{\text{SA}} \;=\; -\log P_\theta(\mathbf{Y} \mid \mathbf{X'})
\;=\; -\log \sum_{\mathbf{A} \in \Gamma} P_\theta\bigl(\mathbf{Y}, \mathbf{A} \,\mid\, \mathbf{X'}\bigr),
\]
where \(\mathbf{X'}\) is the input sequence (with enhanced context), \(\mathbf{\theta}\) denotes the model parameters, and \(P_\theta(\mathbf{Y},\mathbf{A} \mid \mathbf{X'})\) represents the probability of generating the target sequence \(\mathbf{Y}\) through a specific alignment \(\mathbf{A}\).

This objective requires summing over all possible alignment paths \(\mathbf{A} \in \Gamma\), which can be computationally expensive. To address this, we employ a dynamic programming (DP) algorithm that efficiently computes this sum with a complexity of \(\mathcal{O}(M)\), where \(M\) is the number of all masked tokens. 

More specifically, we adopted a dynamic programming algorithm similar to that used in DA-Transformer \citep{huang2022directed}. In this DP scheme, we define \(f_{i,u}\) as the cumulative probability of all partial paths ending at state \(u\) (a node in the DAG) that have generated the first \(i\) tokens of \(\mathbf{Y}\). Formally, \(u\) indexes the states in our DAG in a manner that respects the acyclic property (i.e., we only move forward in the state sequence), and \(i\) ranges over the positions of the target sequence \(\mathbf{Y}\).

The DP recursion works by summing over all valid predecessors \(v\) of \(u\) (where \(v < u\) in the DAG):

\[
f_{i,u} \;=\; \sum_{v \,<\, u} f_{i-1, v} \;\times\; \mathbf{E}_{v,u} \;\times\; \mathbf{P}_{u}(y_i),
\]

where \(\mathbf{E}_{v,u}\) is the transition score from state \(v\) to state \(u\) (derived from our transition matrix), \(\mathbf{P}_{u}(y_i)\) is the probability of state \(u\) predicting the \(i\)-th target token \(y_i\). 

By computing \( f_{i,u} \) across all \( i \) and \( u \), we eventually obtain \( f_{M, L} \), where \( M \) is the number of masked tokens in the target sequence \( \mathbf{Y} \), and \( L \) refers to the final state index. The final training objective is:

\[
\mathcal{L}_{\text{SA}}
\;=\;
-\,\log f_{M,\, L}.
\]

This dynamic programming approach reduces the time complexity of the alignment problem to \(\mathcal{O}(M \times L^2)\), offering a significant improvement over exhaustive path enumeration. In practice, the computation can be further optimized to \(\mathcal{O}(M)\) by leveraging parallelized operations provided by PyTorch \citep{paszke2019pytorch}, making the method highly efficient and suitable for large-scale training. A detailed analysis of the DP algorithm's efficiency is provided in Appendix~\ref{dp_cuda}.

Through this approach, \method{} effectively explores over all possible alignments between the expanded states and target tokens,  leveraging both the token probability distributions \(\mathbf{P}\) and the transition matrix (the edges in DAG) \(\mathbf{E}\)  to reconstruct the missing semantics during pre-training.

\section{Experimental Results}
\label{results_discussions}

We evaluate the performance of the \method{} model in both text modeling and SMILES modeling tasks. SMILES is a sequential representation of molecular information, and a more detailed explanation can be found in Appendix \ref{details_of_smiles}.

We pre-train \method{}\footnote{Code is released at \href{https://github.com/zhengkangjie/ExLM}{https://github.com/zhengkangjie/ExLM}.} on data of text and SMILES separately. Then we fine-tune and evaluate \method~across diverse benchmarks and verify the contribution of each component through ablation studies. Finally, a visualization analysis is included to explain the advantages of unified modeling.

\begin{table*}[t]
\centering
\small 
\caption{
The overall results on the GLUE and SQuAD 2.0 development sets (medians over five random seeds). Results not available in pervious research are marked with ``--''.  The ``MEAN'' column contains the averaged results across the eight GLUE tasks. 
}
\resizebox{\textwidth}{!}{
\begin{tabular}{c*{9}{c}cc}
\toprule
\multirow{2}{*}{\textbf{Model}} & \multicolumn{9}{c}{\textbf{GLUE}} & \multicolumn{2}{c}{\textbf{SQuAD 2.0}} \\ 
\cmidrule(lr){2-10}\cmidrule(lr){11-12}
& \textbf{MNLI-(m/mm)} & \textbf{QQP} & \textbf{QNLI} & \textbf{SST-2} & \textbf{CoLA} & \textbf{RTE} & \textbf{MRPC} & \textbf{STS-B} & \textbf{MEAN} &
\textbf{EM} & \textbf{F1}\\
\midrule
BERT
& 84.5/- & 91.3 & 91.7 & 93.2 & 58.9 & 68.6 & 87.3 & \underline{89.5} & 83.1 &  73.7 & 76.3 \\
ALBERT
& 81.6/- & -- & -- & 90.3 & -- & -- & -- & -- & -- & 77.1 & 80.0 \\
XLNet 
& 85.8/85.4 & -- & -- & 92.7 & -- &--&--&--&--& 78.5 & 81.3 \\
UniLMv2 
& 86.1/86.1 & -- & -- & 93.2 & -- & -- & -- & -- & -- & \underline{80.9} & \underline{83.6} \\
TUPE 
& \underline{86.2}/\underline{86.2} & 91.3 & 92.2 & 93.3 & 63.6 & 73.6 & \textbf{89.9} & 89.2 & 84.9 & -- & --\\ 
RoBERTa$^*$  
& 85.9/85.8 & \underline{91.6} & \underline{92.3} & \underline{93.7} & \underline{64.3} & \underline{75.5} & 88.7 & \underline{89.5} & \underline{85.2} & 78.3 & 81.5 \\
\midrule 
 \rowcolor{mygray} \method{}
& \textbf{86.9/86.7} & \textbf{92.0} & \textbf{93.1} & \textbf{93.9} & \textbf{64.6} & \textbf{78.8} & \underline{89.6} & \textbf{90.5} & \textbf{86.2} & \textbf{82.0} & \textbf{84.6} \\ 
   \rowcolor{mygray}  \method{}$_{\rm LARGE}$
& 87.8/87.5 & 92.2 & 93.8 & 94.5 & 65.3 & 79.1 & 90.4 & 91.2 & 86.9 & 82.6 & 85.0 \\ 

\bottomrule
\end{tabular}
}
\label{tab:main_res}
\end{table*}

\subsection{Results on SMILES Representation Learning}
\label{protein_molecule_tasks}

\paragraph{Pre-training.} For SMILES pre-training, we use the large-scale molecular dataset provided by \citet{zhou2023uni}, which includes SMILES information for $19$ million molecules. We tokenize SMILES sequences with the regular expression from \citet{schwaller2018found}. The pre-training hyperparameters can be found in Appendix \ref{sec::pretrain_configuration}.

\paragraph{Fine-tuning.} 
For fine-tuning, we employ the widely-recognized MoleculeNet benchmark \citep{wu2018moleculenet}. We follow the same data split as used by \citet{zhou2023uni}. Details of the fine-tuning datasets and baselines can be found in Appendix \ref{sec::ft_datasets}. We fine-tune the \method{} model on downstream task datasets using three different random seeds and reported the average performance of the model.

\paragraph{Results.} 
As show in Table \ref{table::classification}, \method{} achieves the best performance among all baseline models on $5$ out of $7$ molecular property classification tasks and closely matches the best baseline models on the remaining two tasks. A noteworthy result is that \method{} significantly outperforms the MLM pre-trained with the same model architecture, pre-training hyperparameters, and pre-training data—namely, the SMILES-BERT model. This further underscores the advantages of \method{} over traditional MLM pre-training tasks. \method{} also demonstrates the strongest average performance across all 7 tasks, indicating that it outperforms other baseline models in these prediction tasks overall.

\subsection{Results on Textual Representation Learning}
\label{text_tasks}

\paragraph{Pre-training.} 
For textual pre-training, we adopt the English Wikipedia and BookCorpus datasets \citep{devlin2018bert} as the pre-training dataset. The model size of \method{} is consistent with BERT base \citep{devlin2018bert}. We also train \method{}$_{\rm LARGE}$ with the same size as BERT large. For more details about pre-training settings, please see Appendix \ref{sec::text_pretrain_configuration}.

\paragraph{Fine-tuning.} 
We evaluate the \method{} model using the the GLUE~\citep{wang2018glue} and SQuAD 2.0~\citep{Rajpurkar2018KnowWY} benchmarks. Detailed of the GLUE benchmark are provided in Appendix~\ref{app:glue}. For fine-tuning, we follow the standard procedures used in BERT~\citep{devlin2018bert} and RoBERTa~\citep{liu2019roberta}. We also provide the hyperparameter search space for fine-tuning in Appendix~\ref{app:hyper}. Consistent with previous studies~\citep{liu2019roberta}, all reported fine-tuning results represent the median values obtained from five different random seeds across both GLUE and SQuAD benchmarks.

\paragraph{Results.} 
As shown in Table \ref{tab:main_res}, we evaluate \method{} on the GLUE and SQuAD 2.0 development sets. \method{} achieves the best performance on SQuAD 2.0 and 7 GLUE tasks, and also performs very closely to the best baseline model on the remaining one GLUE task. Furthermore, \method{} demonstrates significantly higher average performance compared to other baseline models. Notably, RoBERTa in the table uses the same pre-training data and settings as \method{}, yet \method{} shows significant improvements. This demonstrates the superior performance and effectiveness of \method{}. \method{}$_{\rm LARGE}$ also achieves obvious performance improvements compared to \method{} due to the increased scale.

\begin{table}[!t]
\centering
\scriptsize
\caption{
\textbf{Ablation studies.} We assess the effectiveness of 2D RoPE and the transition matrix in \method{} and verify its efficiency. 
}
\begin{tabular}{c|cccc|c}
  \toprule
  Method  & MNLI $\uparrow$ & QNLI$\uparrow$ & QQP$\uparrow$ & RTE$\uparrow$ & Avg  $\uparrow$\\
  \midrule
  
  \method{} w/o 2D RoPE & 84.6 & 91.1 & 91.3 & 56.7 & 80.9 \\
  \midrule
 \method{} w/o Transitions  & 83.8 & 90.9 & 91.1 & 55.6 & 80.4 \\
  \midrule 
  Vanilla MLM  & 83.6 & 90.0 & 90.3 & 54.7 & 79.6\\
  Vanilla MLM++ & 84.4 & 91.2 & 90.6 & 56.3 & 80.7\\
  \midrule 
  \method{}  & \textbf{85.1} & \textbf{91.4} & \textbf{91.3} & \textbf{57.6} & \textbf{81.4} \\
  \bottomrule
\end{tabular}
\label{table::ablations}
\end{table}


\subsection{Analytical Experiments of \method{}}
\label{analysis_exmlm}
We validate \method{}'s effectiveness through ablation studies and explore the impact of expanded states and mask ratios on its performance. A case study also demonstrates its good ability to capture rich semantic information. Efficiency and entropy analysis of \method{} are provided in Appendix \ref{app:more_analysis_exmlm}.

\paragraph{Ablation studies on \method{}.}
We perform ablation studies on the 2D RoPE and transition matrix in \method{}, as shown in Table \ref{table::ablations}. The results reveal that removing the 2D RoPE (\method{} w/o 2D RoPE) or the transition matrix (\method{} w/o Transitions) causes a significant performance drop, with the transition matrix having a larger impact. This demonstrates the importance of the transition matrix in capturing semantic dependencies. Despite the performance decline from removing the transition matrix or the 2D RoPE, the expanded states still enable the model to capture semantic information more effectively, outperforming Vanilla MLM. To assess \method{}'s efficiency, we train an MLM with the same training cost, Vanilla MLM++, and show in Table \ref{table::ablations} that \method{} performs better, highlighting that \method{} can capture semantic information more efficiently than MLM.
\begin{table}[h]
\centering
\scriptsize
\caption{
\textbf{Performance comparison of MLM and \method{} with different mask ratios $p$ and numbers of expanded states \(k\).} \method{} demonstrates stronger modeling capability for contexts with a high mask ratios compared with MLM, and increasing \(k\) further improves its performance under high mask ratios inputs.
}
\begin{tabular}{c|c|cccc|c}
  \toprule
  $p$ & Method  & MNLI $\uparrow$ & QNLI$\uparrow$ & QQP$\uparrow$ & RTE$\uparrow$ & Avg  \\
  \midrule
    $15\%$ & Vanilla MLM & 83.6 & 90.0 & 90.3 & 54.7 & 79.6\\
  \midrule
  \multirow{3}{*}{$15\%$} & \method{}-k=$2$ & 84.6 & 91.3 & 91.1 & 56.7 & 80.9 \\
   & \method{}-k=$4$ & \textbf{85.1} & \textbf{91.4} & \textbf{91.3} & \textbf{57.6} & \textbf{81.4} \\
  & \method{}-k=$8$ & 84.4 & 91.0 & 90.9 & 56.9 & 80.8 \\
  \midrule
  $38.7\%$ & Vanilla MLM & 83.3 & 90.0 & 90.2 & 53.3 & 79.2 \\
  \midrule
  \multirow{3}{*}{$38.7\%$} & \method{}-k=$2$ & 84.1 & 91.4 & 91.0 & 55.4 & 80.5 \\
   & \method{}-k=$4$ & 84.9 & 91.6 & 91.2 & 57.0 & 81.2 \\
  & \method{}-k=$8$ & 84.3 & 90.9 & 90.7 & 56.8 & 80.7 \\
  \bottomrule
\end{tabular}
\label{table::impact_ratio}
\end{table}
\paragraph{The impact of $k$ and $p$.}
We explore the impact of the number of expanded states $k$ and mask ratios $p$ on \method{}'s performance, with results shown in Table \ref{table::impact_ratio}. As $k$ increases, performance improves due to enhanced semantic modeling ability. However, when $k$ becomes too large (e.g., $k=8$), performance slightly declines due to the excessive expanded states of \mt{} tokens in the input context and redundant nodes in the DAG. Thus, with a mask ratio of $15\%$, $k=4$ proves to be the optimal choice. We further verify the impact of the mask ratio \( p \) on the performance of \method{}, with the results shown in Table \ref{table::impact_ratio}. These results show that a higher mask ratio \( p \) has less impact on \method{}'s performance compared to MLM, highlighting the enhanced semantic modeling capability of \method{}. Furthermore, \method{} needs more expanded states to model the multimodal and ambiguous context induced by a higher mask ratio. Thus, increasing \( k \) appropriately at higher mask ratios enhances its performance. However, an excessively high mask ratio and too many expanded states (a higher \(k\)) can result in a large number of expanded states of \mt{} tokens in the input context, limiting further performance gains of \method{}.


\begin{figure}[t]
\centering
	\includegraphics[width=1.0\linewidth]{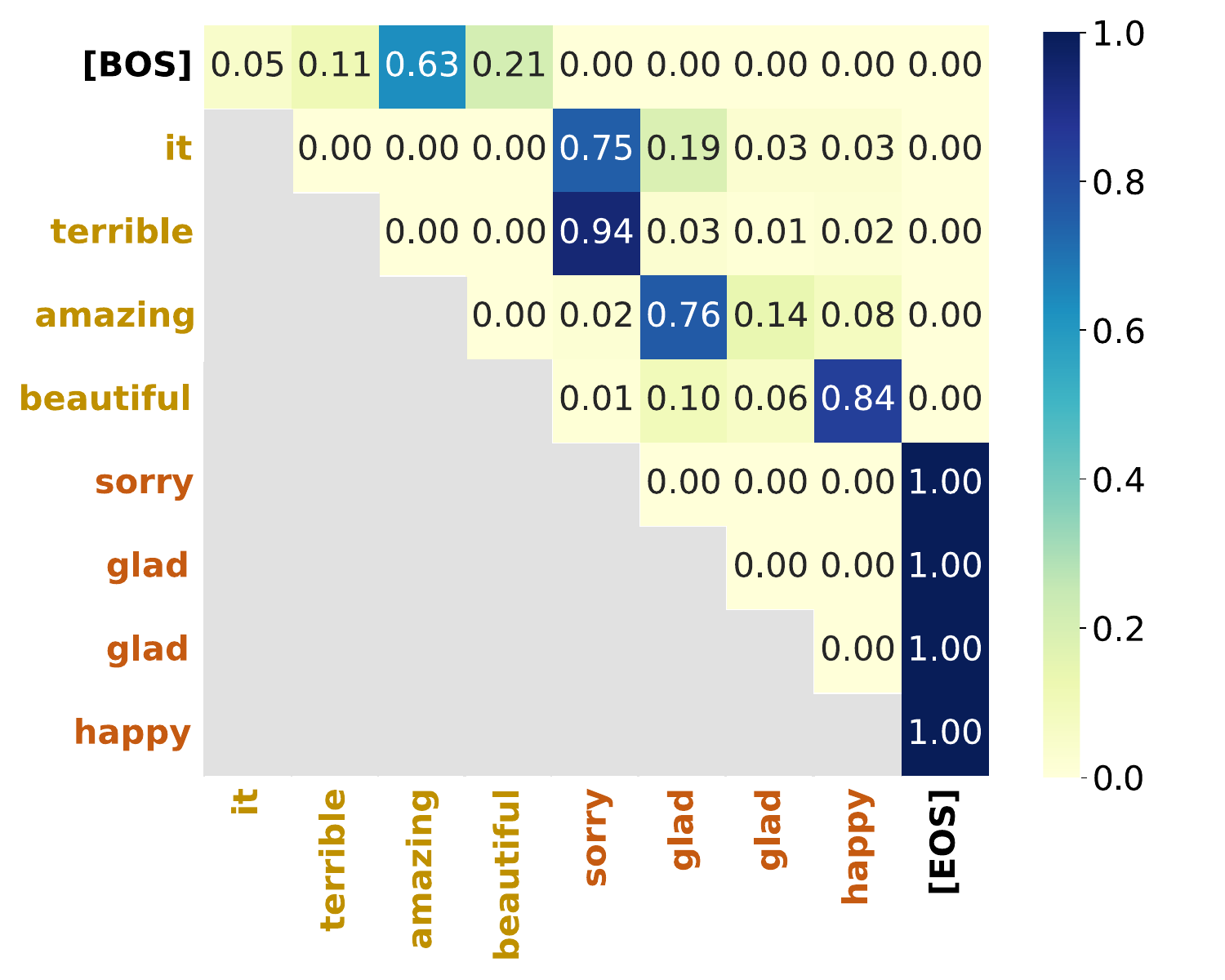}
	\caption {\textbf{Case study.} We visualize the model's predictions when the input is ``\textit{This is \mt{}, and I’m very \mt{} to see this}." ($k=4$). The \textcolor[HTML]{BF9000}{\textbf{yellow nodes}} correspond to the expanded states of the first \mt{} token, while the \textcolor[HTML]{C55A11}{\textbf{brown nodes}} correspond to those of the second \mt{} token. The weights in the figure represent the transition probability between two nodes in the DAG, and the axis labels show the top-1 predicted word for each node. }
\label{fig::case}
\end{figure}

\paragraph{Case Study.}
We conduct a case study to evaluate the model's performance on specific inputs, as shown in Figure \ref{fig::case}. The results demonstrate that the model effectively handles multimodal semantic information, capturing four possible semantic choices for each \mt{} token. The transition matrix models the relationships between these choices, for example, when the first \mt{} token is \textit{amazing}, the second is more likely to be \textit{glad}, and when the first is \textit{terrible}, the second tends to be \textit{sorry}. A more detailed analysis can be found in Appendix \ref{app:case_study}. These results show that \method{} can capture rich contextual information. 

\section{Conclusions}
\label{conclusions}
In this paper, we analyze the impact of the semantics corruption caused by \mt{} tokens on MLMs, showing that it has a more significant impact than the unreal token problem, offering a new perspective on understanding MLMs. Based on our analysis, we propose \method{}, which enhances the model’s semantic modeling ability through two key designs: States Expansion and Dependency Capture. These designs reduce the negative impact of the semantic multimodality on the model. We demonstrate \method{}'s strong performance in both text and SMILES modeling scenarios. Ablation studies and case study also validate its effectiveness and efficiency.

\section*{Acknowledgements}
This paper is partially supported by grants from the National Key Research and Development Program of China with Grant No. 2023YFC3341203 and the National Natural Science Foundation of China (NSFC Grant Number 62276002). We would like to thank Jie Zhu from University of Oxford and Liang Ding from the JD Explore Academy for their insightful discussions on the project. We also thank all other members from Dlib in PKU for their valuable feedback given during the internal discussions. 

\section*{Impact Statement}
This paper is the first work to systematically study the impact of \mt{} tokens on MLM from the perspectives of unreal tokens and semantics corruption, and proposes effective solutions to these problems. These analytical results provide valuable insights for designing better language representation learning models. Furthermore, MLM-based sequence representation learning models have already achieved significant success in various domains, including text, images, video, and scientific data. These models have also shown promising results in applications such as recommendation systems and protein design. Therefore, the methods proposed in this paper have the potential to drive further advancements in these fields, and the approach also holds significant application potential. However, we also acknowledge that this work inherits the potential negative impacts of existing MLMs, such as the possibility of being used to generate false information or design and manufacture molecules with biological hazards.

\bibliography{example_paper}
\bibliographystyle{icml2025}

\newpage
\appendix
\onecolumn

\begingroup
\hypersetup{colorlinks=false, linkcolor=red}
\hypersetup{pdfborder={0 0 0}}
\renewcommand\ptctitle{}

\part{Appendix} 
\parttoc 


\endgroup


\section{Related Works}
\subsection{Language Models for Representation Learning.}
\label{lm_pretraining}
\paragraph{MLM for Representation Learning.}Currently, language models serve as an important role in the self-supervised representation learning field and achieve excellent performance across a wide range of tasks. The ELMo model \citep{peters-etal-2018-deep} first introduces a self-supervised language representation learning model based on bidirectional LSTM \citep{hochreiter1997long}. Later, the pre-trained masked language model based on the transformer architecture, BERT \citep{devlin2018bert}, is proposed in the field of text modeling. BERT uses a bidirectional transformer encoder to extract meaningful semantic representations from large amounts of text data and applies them to various downstream tasks such as text understanding \citep{devlin2018bert} and text correction \citep{zhang2020spelling, dai2022whole,zheng2023towards}. With the help of the transformer architecture \citep{vaswani2017attention} and the MLM training approach, this model achieves great success in the text domain. Following this, a large number of follow-up works aim to improve these bidirectional encoder-based masked language models \citep{liu2019roberta, ke2020rethinking, shin2020fast, namazifar2021warped, du2021glm, fu2022contextual, wettig2022should, meng2022pretraining, meng2023representation, he2022diffusionbert}. Moreover, MLM also succeeds in various representation learning tasks, including images \citep{bao2021beit, xie2022simmim, he2022masked}, videos \citep{tong2022videomae, wang2022bevt}, codes \citep{feng2020codebert, guo2020graphcodebert, he2021debertav3}, small molecules\citep{wang2019smiles, chithrananda2020chemberta, ross2022large, pan2023large, feng2024bioactivity, yang2024mol, zheng2024smi}, proteins \citep{elnaggar2021prottrans, lin2022language, lin2023evolutionary, zhengesm, hayes2025simulating}, and single-cell data \citep{yang2022scbert, theodoris2023transfer}. 

\paragraph{Other Methods \& Summary.} In addition to language models trained with the MLM pre-training approach, there are other types of language models used for representation learning tasks, such as autoregressive language models \citep{wang2023improving, jiang2023scaling, springer2024repetition, ren2024representation, li2024bellm}, diffusion language models \citep{wang2024diffusion, wang2024dplm}, and unified language models \citep{dong2019unified, bao2020unilmv2, du2021glm}. These models have demonstrated their effectiveness in representation learning across various tasks. However, overall, MLM is one of the most commonly used approaches for constructing language models for representation learning tasks and is a typical representative of this type of model. \citet{zhong2023can} also points out that the MLM performs better than LLM in handling paraphrase and similarity tasks. Given the widespread impact and application of MLM, studying the corrupted semantics problem in MLM is of great significance, and it may also play a positive role in advancing the future development of MLM in different fields.

\subsection{Studies of \mt{} in MLM pre-training.}
\label{mask_pretraining}
Due to the widespread use of MLM and the important role of the \mt{} tokens in the MLM training process, several previous works have studied the impact of the \mt{} tokens on MLM. \citet{clark2020electra} proposed the ELECTRA model, an improvement on MLM, where the \mt{} token is not included in the context during pre-training. This approach helps address the inconsistency between pre-training and downstream task fine-tuning caused by \mt{} unreal tokens. \citet{liao2022mask} further studied the impact of the \mt{} unreal tokens on MLM models and pointed out that removing \mt{} tokens from the model's shallow representations during training does not affect MLM performance. Meanwhile, \citet{meng2023representation} explored the representation deficiency problem caused by \mt{} unreal tokens from a theoretical perspective and proposed a MAE-based language model training approach. This approach improves the model's representation learning ability by removing the \mt{} tokens from the model input to avoid the unreal tokens problem. Additionally, \citet{wettig2022should} and \citet{liao2022mask} also explored the impact of higher mask ratio on MLM performance. While these works have studied the effects of the \mt{} tokens, they focus only on the unreal tokens aspect and lack an investigation into the model's behavior from the perspective of semantic corruption. In contrast, \citet{zhong2023revisiting} points out that dropping \mt{} tokens causes semantic loss, affecting performance on semantic-intense tasks like RTE \citep{RTE-1}. This finding further highlights the negative impact of token dropping on the model's semantic modeling capability, emphasizing the importance of addressing the semantic loss problem. However, this work only analyzes the additional semantic loss caused by the token dropping process compared to the token masking process, and does not analyze the semantic loss caused by the token masking process itself. Therefore, its conclusion can not generalize to MLMs. This work aims to fill this gap by analyzing MLM's behavior from the perspective of both the semantic corruption problem and the unreal tokens problem.

\clearpage

\section{Hyper-Parameter Configuration for \textit{Repeated MLM} Experiments}
\label{sec::analyse_configuration}

\subsection{Pre-training Configuration}
In the \textit{Repeated MLM} experiment, we train a series of MLMs with different $p$ and $k$ parameters (a total of 14 sets of models). When $k$ is large, the input sequence length of the model increases significantly, which also results in a high training cost. To accelerate the training process, we use a larger batch size and a larger learning rate to help the model converge more quickly, and we also reduce the number of training steps to lower the overall training cost, ensuring that the training expense is manageable. The model size used in this part of the experiment is the same as BERT base \citep{devlin2018bert}. Specifically, the model has $12$ stacked Transformer layers, each with $12$ attention heads. The model dimension and feedforward dimension of each Transformer layer are $768$ and $3{,}072$, respectively. The total number of parameters in the model is $128$M. Furthermore, to ensure that the increase in $k$ does not result in an excessively long sequence that the model cannot process, the model's \textit{Max sequence length} parameter (the maximum length of the position embedding) increases with $k$, with a specific value of $512*k$. And we set the learning rate as $0.002$ and warmup steps as $5$K. The total training steps are $50$K and each batch has $4096$ samples at maximum. 

To ensure the comparability of the results, all other pre-training hyperparameters in the MLM across different experimental groups are exactly the same, except for the \textit{Max sequence length} parameter. In addition to the \textit{Repeated MLM} experiment, we also used the same pre-training parameters in the ablation experiments to reduce pre-training costs. For more pre-training hyper-parameters, please refer to Table \ref{analyse_table_hyper}.

\begin{table}[ht]
\footnotesize
\color{black}{
\caption{Hyper-parameters for the \textit{Repeated MLM} experiments.}
\label{analyse_table_hyper}

\begin{center}
\begin{tabular}{ccc}
\toprule
\multicolumn{1}{c}{\multirow{1}{*}{Hyper-parameters}} &\multicolumn{1}{c}{\multirow{1}{*}{Value}}\\
\midrule
Learning rate  & 2e-3 \\
LR scheduler  & polynomial\_decay \\
Warmup updates & $5$K \\
Max updates & $50$K \\
Batch size & $4,096$ \\
FFN dropout & $0.1$ \\
Attention dropout & $0.1$ \\
Activation dropout & $0$ \\
Num of layers & $12$ \\
Num of attention heads & $12$ \\
Encoder embedding dim & $768$ \\
Encoder FFN dim & $3,072$ \\
Adam ($\beta_1, \beta_2$) & $(0.9,0.98)$ \\
Mask ratio &  $0.15$ \\
Activation function & GELU \\
Weight Decay & $0.01$ \\
Clip Norm & $0.0$ \\
Max sequence length & $512$*$k$ \\
\bottomrule
\end{tabular}
\end{center}
}
\end{table}
\subsection{Fine-tuning Configuration}
To ensure the comparability of the experiments, the MLMs in different sets of experiments use exactly the same training parameters during the fine-tuning phase (except for the \textit{Max sequence length} parameter, which is set the same as in the pre-training phase). Moreover, to ensure consistency between the pre-training and fine-tuning processes, we repeat each token in the input sequences $k$ times during fine-tuning, using the same method as in the pre-training phase. Each MLM is trained three times with different random seeds on each downstream task, and the average of these three results is taken as the final outcome. The detailed training parameters are provided in Table \ref{tab:analyse_glue_hyper}.

\begin{table*}[h]
    \centering
    \begin{tabular}{lc}
        \toprule
        \textbf{Hyperparameter} & MNLI, QNLI, QQP, RTE  \\
        \midrule
        Peak learning rate & 5e-5 \\
        Batch size &32  \\
        Max epochs & 3 \\
        Warm-Up Proportion & 6\% \\
        Max sequence length & 512*$k$ \\
        Fine-tuning Seeds & \{1, 2, 3\} \\
        \bottomrule
    \end{tabular}
    \caption{Fine-tuning Hyper-Parameter Configuration for \textit{Repeated MLM} Experiments.}
    \label{tab:analyse_glue_hyper}
\end{table*}

\section{Expectation and Variance of the Proportion of Corrupted Semantics }
\label{ev_csp}

To quantify the proportion of corrupted semantics, consider a sequence of \( N \) tokens, and each token is duplicated by \( k \) times, resulting in a total of \( N \times k \) tokens. Each duplicated copy is independently masked with probability \( p \). A token is deemed to have corrupted semantics if all \( k \) of its copies are masked. Formally, the corrupted semantics proportion \( s \) is defined as:

\[
s = \frac{1}{N} \sum_{i=1}^{N} X_i
\]

where \( X_i \) is an indicator variable defined by:

\[
X_i =
\begin{cases}
1, & \text{if all } k \text{ copies of token } i \text{ are masked}, \\
0, & \text{otherwise}.
\end{cases}
\]

Each of these \( k \) copies is independently masked with probability \( p \). To determine if the semantics of token \( i \) are corrupted, we need all \( k \) copies of the token to be masked simultaneously. Thus the probability that all \( k \) copies of token \( i \) are masked is simply the product of the individual masking probabilities. Therefore, the probability that \( X_i = 1 \), meaning that all \( k \) copies are masked and the token’s semantics are corrupted, is given by:

\[
\mathbb{P}(X_i = 1) = p^k.
\]

Since \( X_i \) is an indicator variable that takes the value 1 if the token’s semantics are corrupted, and 0 otherwise, it follows a Bernoulli distribution with parameter \( p^k \), i.e., \( X_i \sim \mathrm{Bernoulli}(p^k) \). Thus the expectation of $X_i$ is $p^k$ and the variance of $X_i$ is $p^k (1 - p^k)$.

Therefore the expectation and variance of \( s \) can be derived as follows:

\[
\mathbb{E}[s] = \mathbb{E}\!\Bigl[\frac{1}{N}\sum_{i=1}^{N} X_i\Bigr] = \frac{1}{N} \sum_{i=1}^{N} \mathbb{E}[X_i] = p^k,
\quad
\mathrm{Var}(s) = \mathrm{Var}\! \Bigl(\frac{1}{N}\sum_{i=1}^{N} X_i\Bigr) = \frac{1}{N^2} \sum_{i=1}^{N} \mathrm{Var}(X_i) = \frac{p^k (1 - p^k)}{N}.
\]

This analysis leverages the independence of masking each token. It proves that when the repetition times are \(k\) and the mask ratio is \(p\), the proportion of corrupted semantics is \(p^k\). Moreover, because the input sequence length $N$ is typically large, the variance of $s$ becomes vanishingly small. Concretely, let’s take:
\[
p = 0.387,\quad k = 2,\quad N = 512.
\]
Then
\[
\mathbb{E}[s] = p^k = 0.387^2 \approx 0.1498,
\]
\[
\mathrm{Var}(s)
= \frac{0.1498\,(1 - 0.1498)}{512}
\approx 2.49 \times 10^{-4},
\]
\[
\sigma_s = \sqrt{\mathrm{Var}(s)} \approx 0.0158.
\]
Thus, even though the expected corrupted semantics proportion is about $15\%$, the standard deviation is only around $1.6\%$, confirming that $s$ is tightly concentrated around $p^k$ when $N$ is large.

\clearpage

\section{A Brief Introduction to SMILES}
\label{details_of_smiles}

SMILES (Simplified Molecular Input Line Entry System) \citep{weininger1988smiles} is a notation system used to represent chemical structures in a text format. It encodes molecular information using a series of characters, where atoms are represented by their chemical symbols (e.g., C for carbon, O for oxygen), bonds by symbols like ``-" (single bond), ``=" (double bond), and ``\#" (triple bond), and branches are indicated with parentheses. For example, the SMILES representation of ethanol is ``CCO", where ``C" stands for carbon atoms and ``O" represents an oxygen atom, with a single bond between them. The notation captures the connectivity between atoms, bond types, and sometimes stereochemistry, making it a compact and easily readable way to represent molecular structures. The principle behind SMILES is that it provides a linear representation of the molecule that can be easily interpreted by both humans and computational systems, facilitating tasks like database searching, molecular simulations, and chemical informatics.

\section{Calculation of Entropy in \textit{Repeated MLM}}
\label{app::ent_details}

To analyze the uncertainty of MLM predictions, we calculate the entropy of the model's predicted distribution $P$ for missing tokens. Entropy is a measure of uncertainty, calculated as:

\[
H[P] = - \sum_{x\in{\rm voc}} P(x) \log P(x),
\]

where \(P(x)\) is the predicted probability of token \(x\) at a given masked position, ${\rm voc}$ is the vocabulary. 

A larger entropy indicates a more dispersed prediction distribution $P$, which also suggests that the model has higher uncertainty.

\subsection{Adjustments for Repeated Tokens} 

In the \textit{Repeated MLM} experiment, some tokens are repeated multiple times, which introduces additional considerations for calculating entropy. Without proper filtering, the calculated entropy would be influenced by tokens that are not fully masked. This means that the entropy would reflect the uncertainty associated with the model’s predictions for partially visible tokens, resulting in an artificially lower entropy that fails to represent the true uncertainty under full masking conditions. To ensures the entropy calculation only considers tokens where all \(k\) copies are masked, we introduce a filtering criterion. Any token with at least one unmasked copy is excluded from the computation. 

For a token that appears \(k\) times and is fully masked, the entropy is determined by averaging the individual entropies of the prediction distributions for each of its masked copies. This approach gives a more consistent and meaningful representation of the model’s uncertainty. Formally, the entropy of a token with \(k\) masked positions is defined as:

\[
H_{\text{token}} = \frac{1}{k} \sum_{j=1}^{k} H[P(x_j)],
\]

where \(H[P(x_j)]\) is the entropy of the prediction distribution at the \(j\)-th repetition of the fully masked token, \(P(x_j)\) represents the model’s predicted probability distribution over the vocabulary at that position. The summation iterates over all \(k\) masked positions, and the resulting average provides the token’s overall entropy.

This method ensures that the entropy metric accurately reflects the model’s uncertainty in the fully masked scenario, without interference from partially visible tokens or the added complexity of repeated instances.

\subsection{Data and Experimental Setup} 

This entropy analysis is conducted on the validation set to ensure that the results reflect the model's behavior on unseen data and avoid any training bias. By focusing on fully masked tokens and calculating their entropy, this approach provides a precise measure of the model's uncertainty in predicting missing semantic information.

\clearpage 
\section{More Results of \textit{Repeated MLM} Experiments}
\label{sec::more_repeat_mlm}

\begin{figure*}[h]
\centering
\subfigure[MNLI-m]{\includegraphics[width=0.49\textwidth]{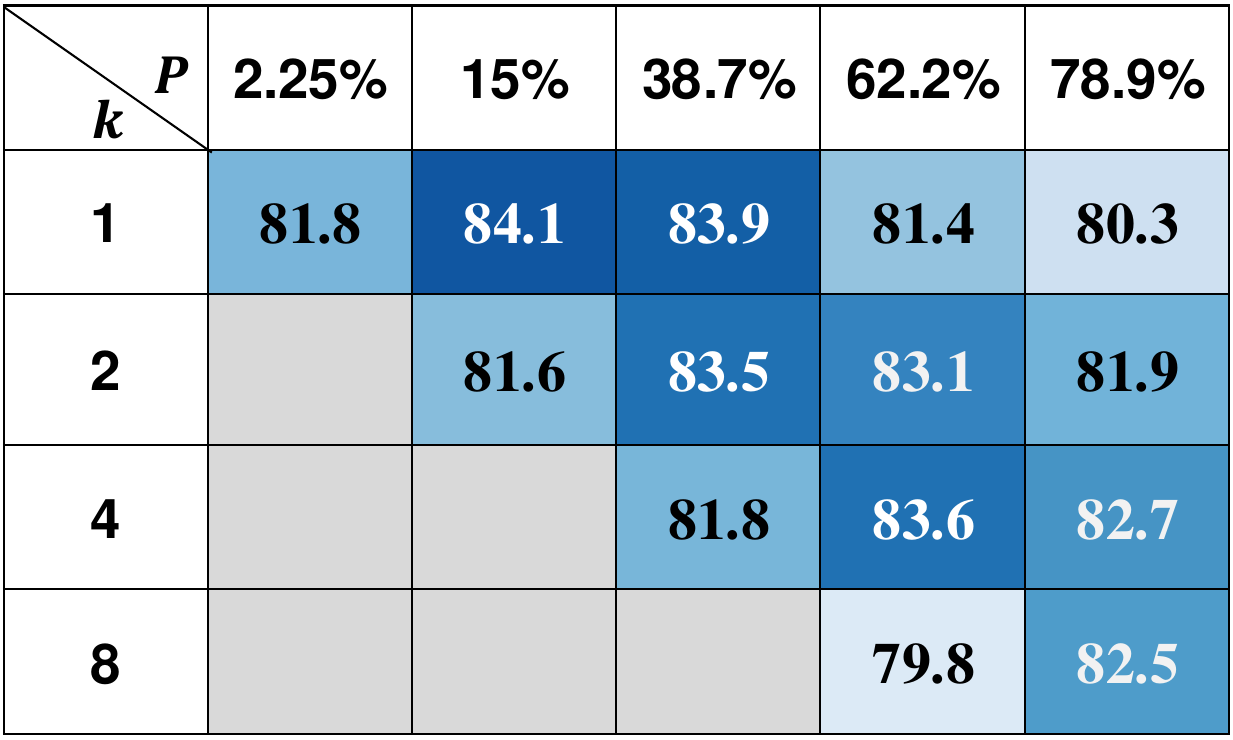}
}
\subfigure[QNLI]{\includegraphics[width=0.49\textwidth]{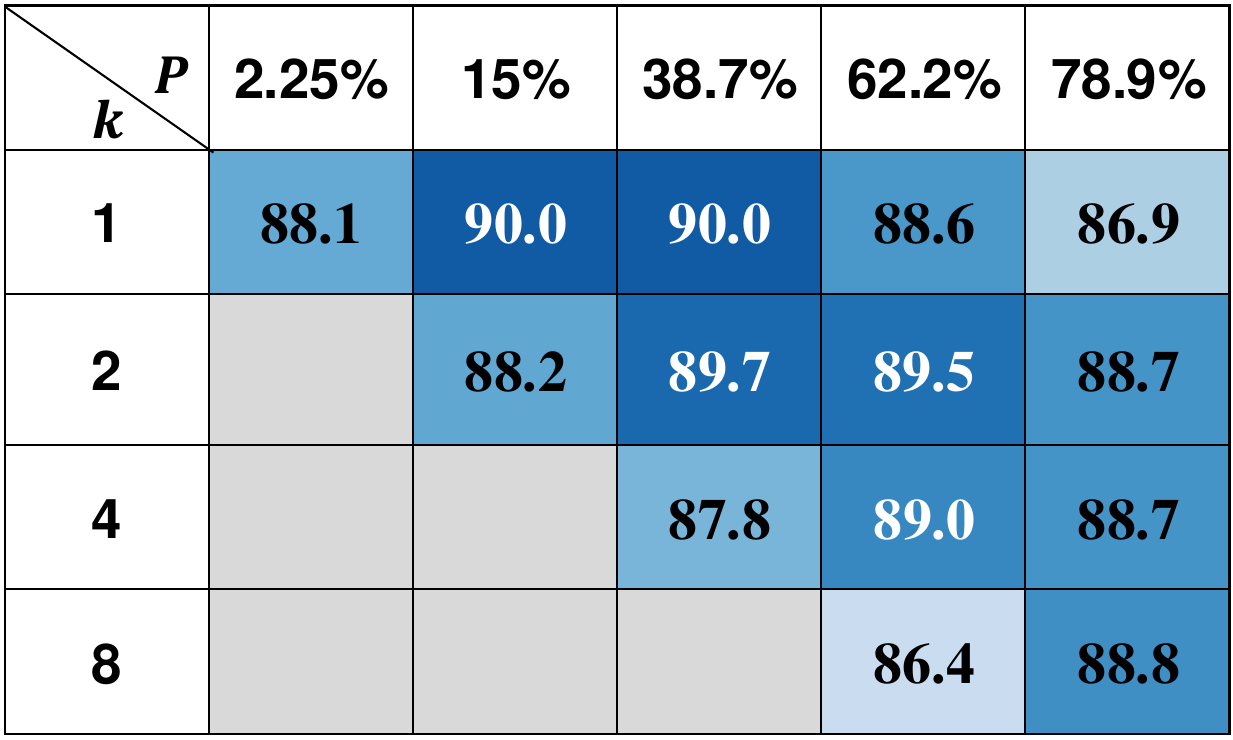}}

\subfigure[QQP]{\includegraphics[width=0.49\textwidth]{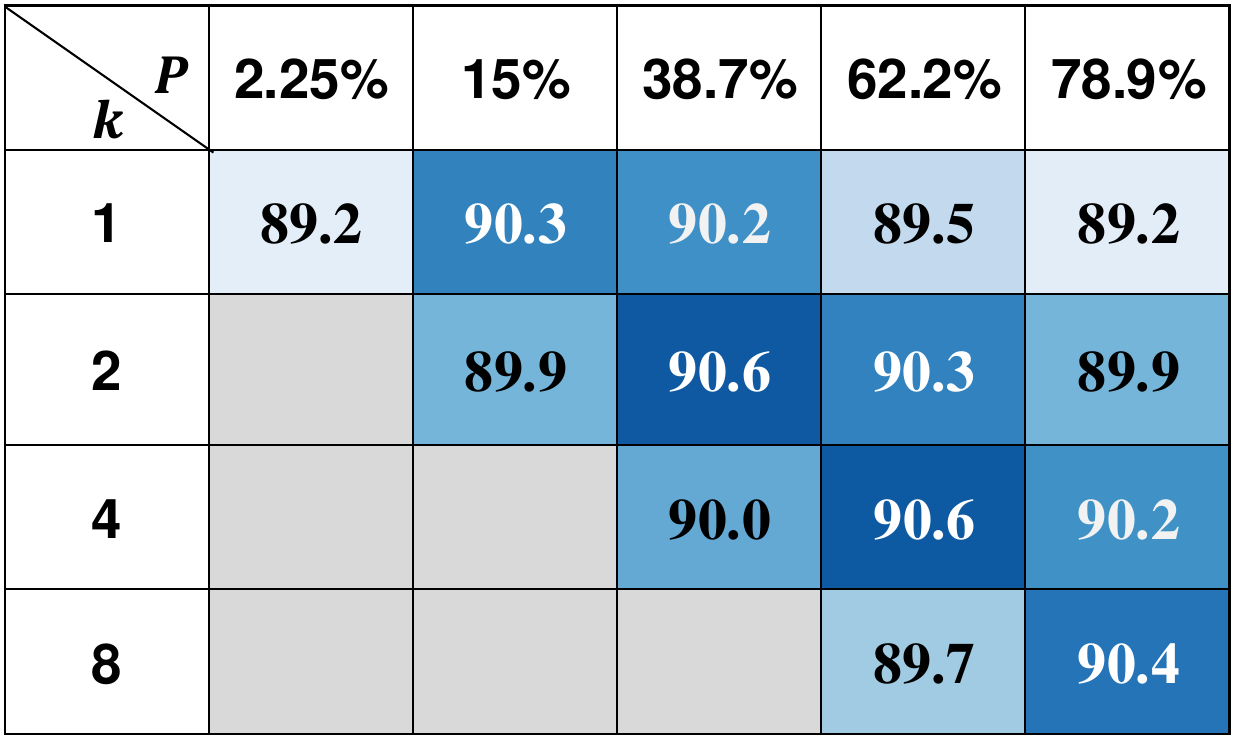}
}
\subfigure[RTE]{\includegraphics[width=0.49\textwidth]{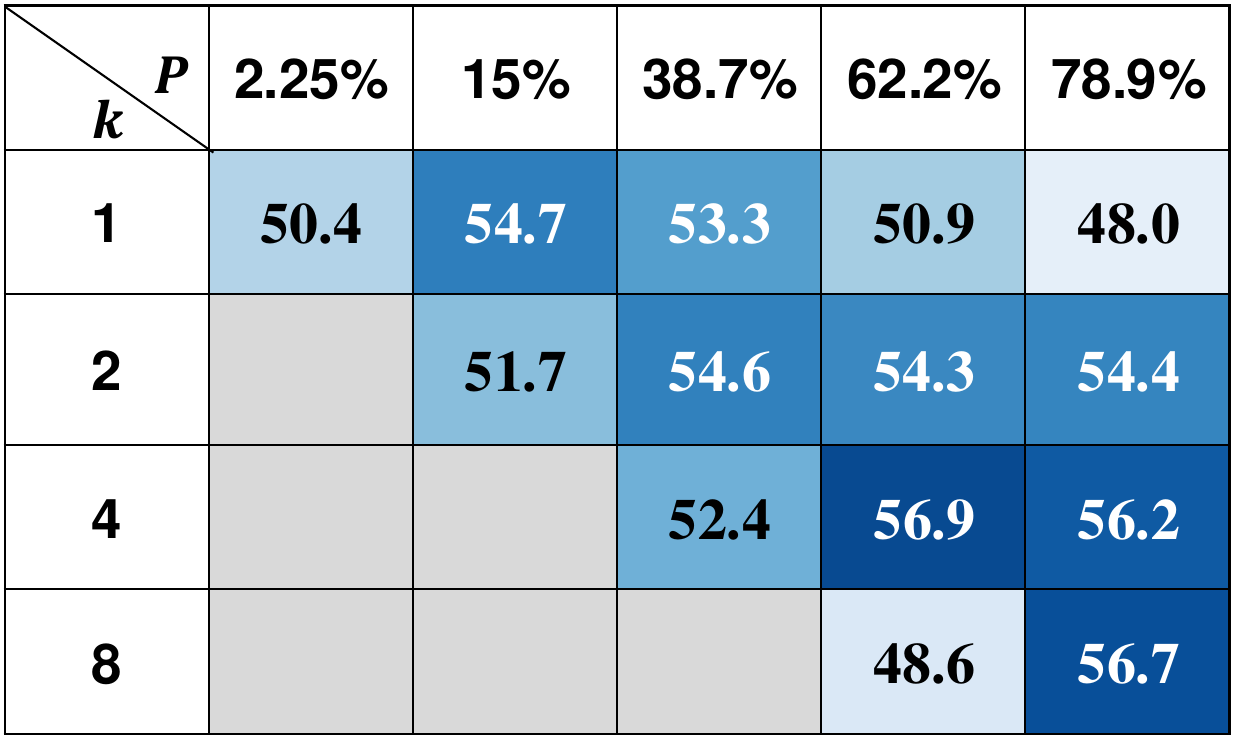}
}

\caption{The results of the \textit{Repeated MLM} experiments on MNLI-m, QNLI, QQP, and RTE tasks. We use accuracy in these four tasks as the metrics. And for all tasks, higher values are better. Similar performance are marked with similar colors.}
\label{fig_more_repeat_mlm}
\end{figure*}

We extend the \textit{Repeated MLM} experiment to a broader range of tasks, with the results presented in Figure \ref{fig_more_repeat_mlm}. From these results, we observe that the conclusions drawn in Section \ref{sec::compare_results_analyse} still hold. Although the exact performance trends vary from task to task, the overall pattern remains clear and consistent: \textbf{the severity of corrupted semantics directly correlates with performance degradation}. 

Specifically, as the intensity of corrupted semantics changes, the model’s performance on these tasks exhibits significant fluctuations. Notably, the magnitude of these changes is substantially greater than the performance variations observed when the mask ratio is altered. This suggests that corrupted semantics plays a more critical role in influencing model performance than the presence of unreal tokens.

\clearpage 

\section{Training Curves in the \textit{Repeated MLM} Experiments}
\label{sec::repeat_mlm_curve}

\begin{figure*}[h]
\centering

\subfigure[Loss and Acc (Mask Ratio=$38.7$\%)]{\includegraphics[width=0.39\textwidth]{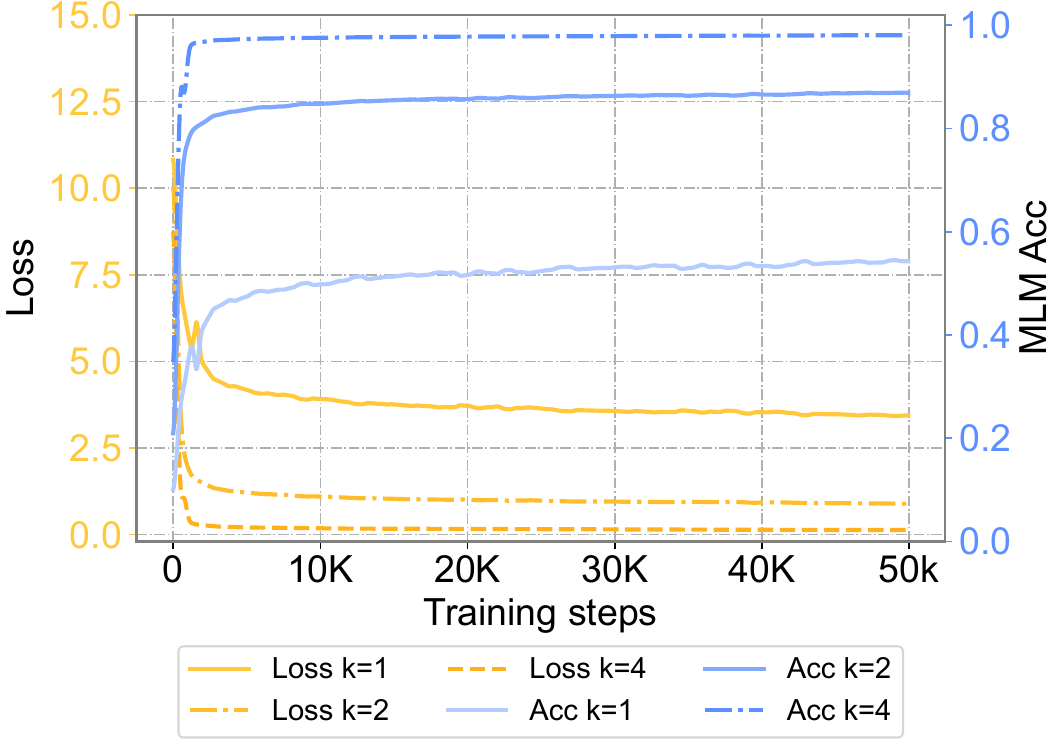}
}
\subfigure[PPL (Mask Ratio=$38.7$\%)]{\includegraphics[width=0.39\textwidth]{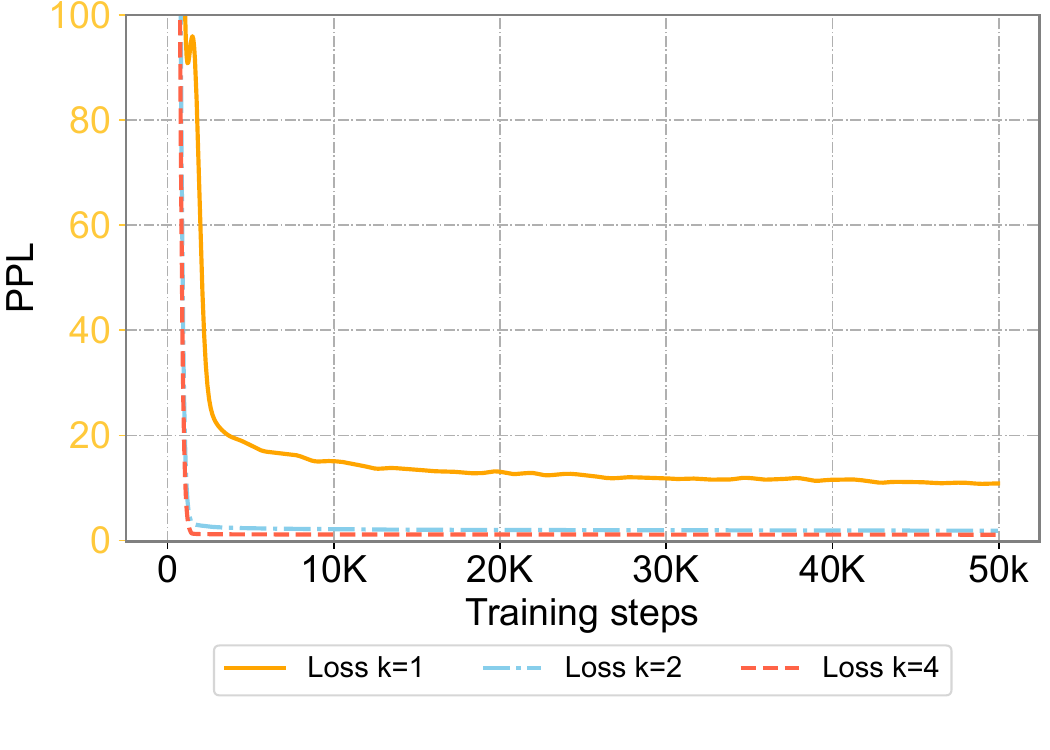}}

\subfigure[Loss and Acc (Mask Ratio=$62.2$\%)]{\includegraphics[width=0.39\textwidth]{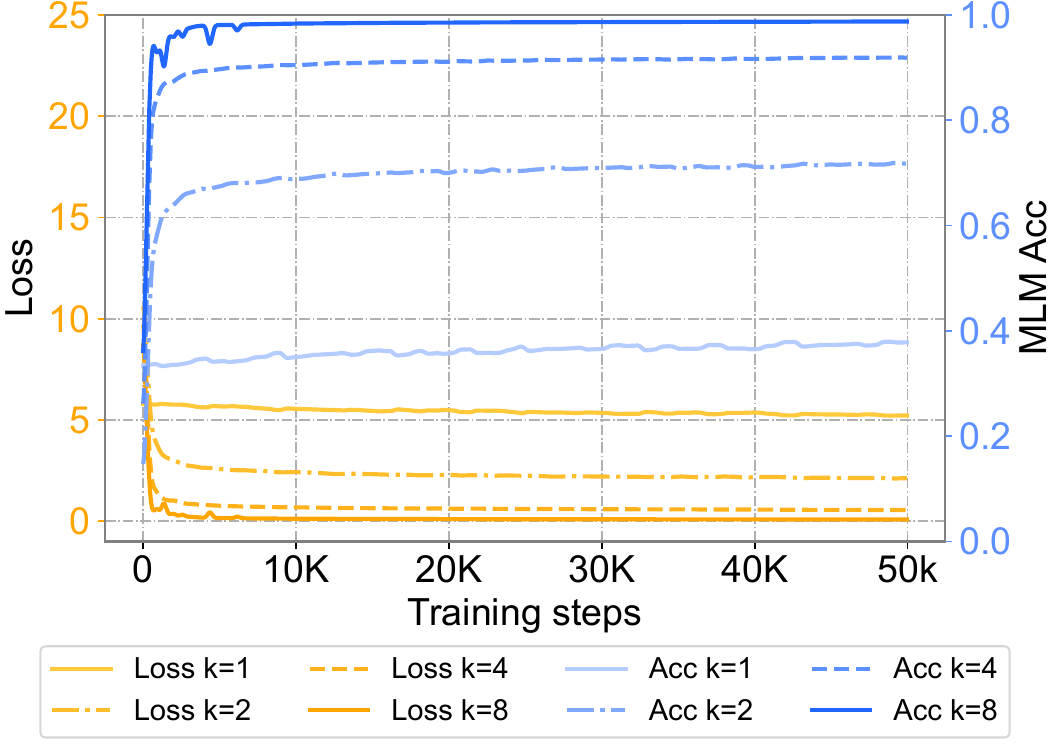}
}
\subfigure[PPL (Mask Ratio=$62.2$\%)]{\includegraphics[width=0.39\textwidth]{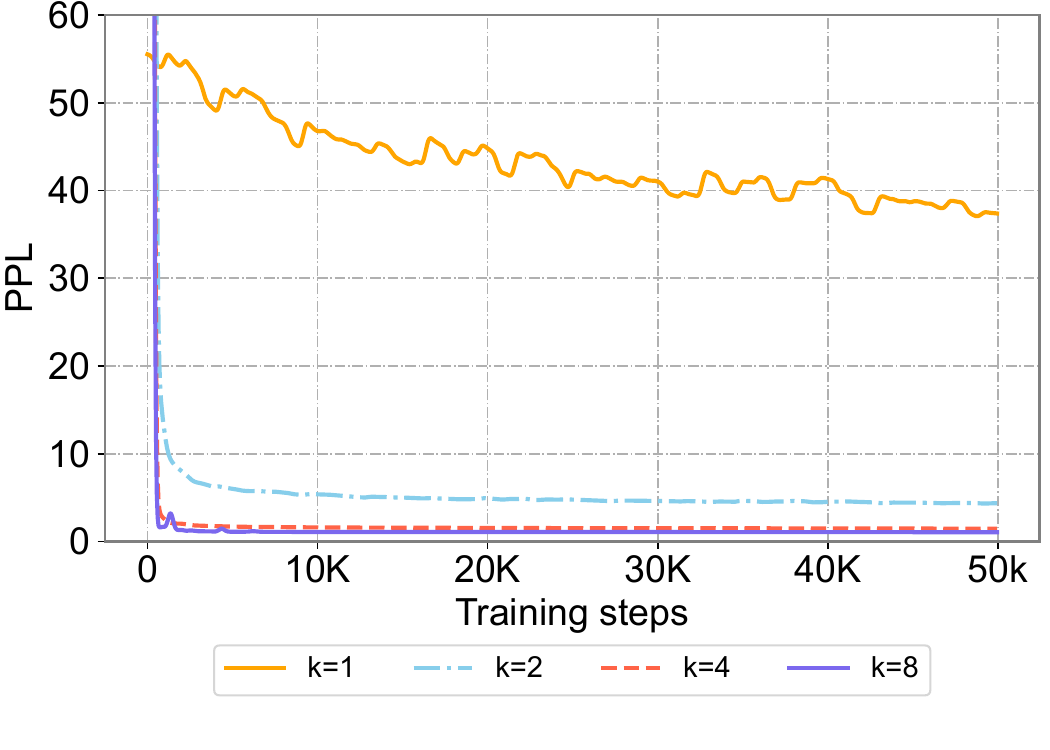}}

\subfigure[Loss and Acc (Mask Ratio=$78.9$\%)]{\includegraphics[width=0.39\textwidth]{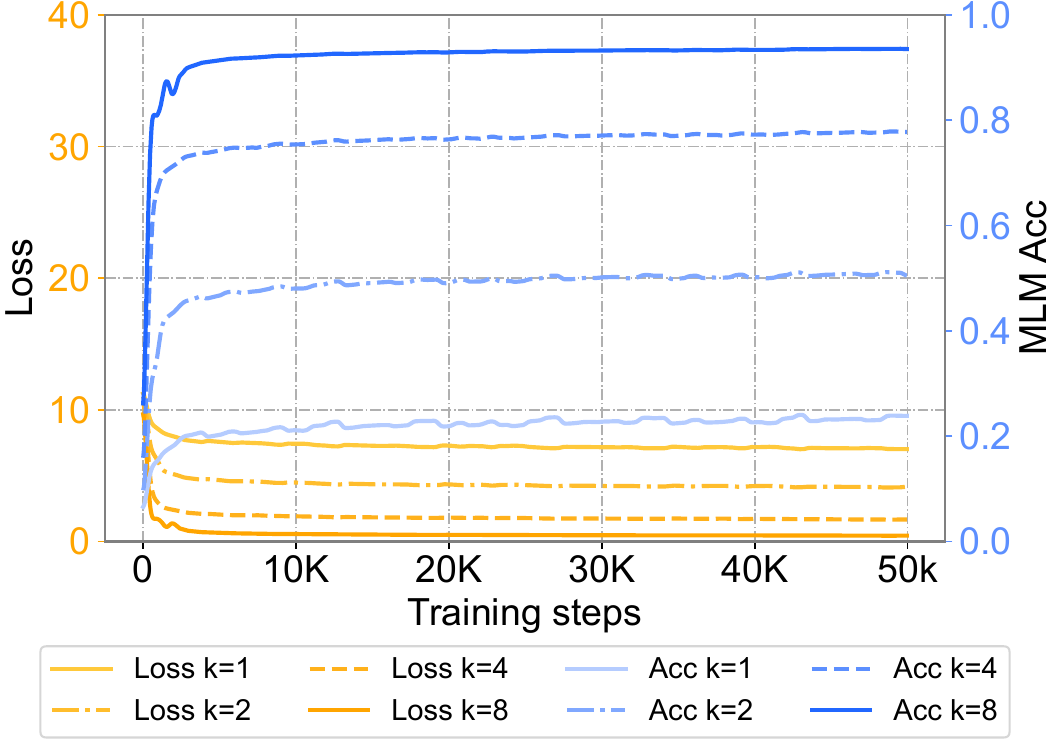}
}
\subfigure[PPL (Mask Ratio=$78.9$\%)]{\includegraphics[width=0.39\textwidth]{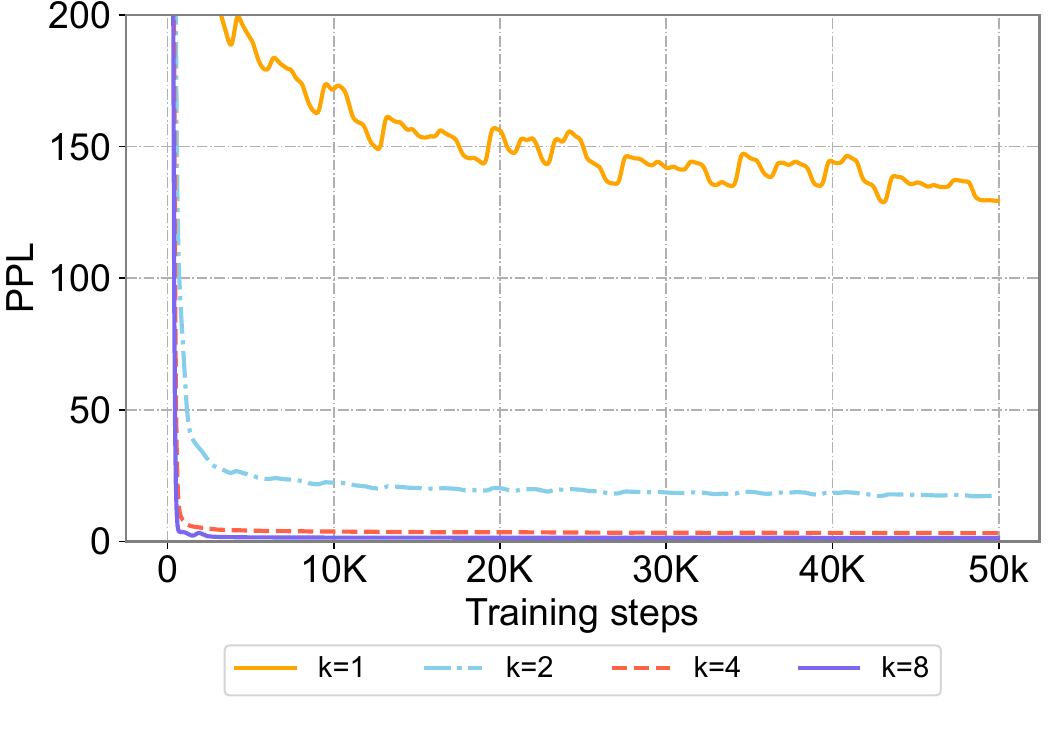}
}

\caption{The training loss, mask prediction accuracy and perplexity (PPL) curves of MLMs in the \textit{Repeated MLM} experiments.}
\label{fig_repeat_mlm_curves}
\end{figure*}

We plot the training curves of the MLMs with different repetition times \( k \) and mask ratios \( p \) in the \textit{Repeated MLM} experiments, as shown in Figure \ref{fig_repeat_mlm_curves}. From this figure, we can observe that when we fix the mask ratio \( p \) and change the parameter \( k \), as \( k \) increases (meaning more redundancy in the input), the intensity of semantics corruption in the context received by the model decreases. As a result, the model exhibits faster convergence, lower loss, lower perplexity and higher mask prediction accuracy during pre-training. This is also a key reason why model performance drops significantly when the semantics corruption in the input context are too low; at this point, the pre-training task becomes too simple, and the model converges quickly without learning meaningful information.

\clearpage

\section{CUDA-Accelerated Dynamic Programming Framework: Details and Efficiency Analysis}
\label{dp_cuda}

Naively summing over all paths \(\mathbf{A} \in \Gamma\) can be prohibitively expensive, since the number of possible paths grows exponentially. To solve this, we use a dynamic programming (DP) approach, which is also adopted by the training algorithms for connectionist temporal classification \citep{graves2006connectionist, saharia-etal-2020-non} and DA-Transformer \citep{huang2022directed,shao2022viterbi,huang2022PDAT, huang-etal-2024-decoding}. We further adopted a CUDA-accelerated dynamic programming algorithm from DA-Transformer \citep{huang2022directed} to efficiently perform state alignment during pre-training. The core of the algorithm utilizes CUDA (Compute Unified Device Architecture) to parallelize the computations involved in dynamic programming for sequence modeling. The primary objective is to efficiently calculate forward (\(\alpha\)) and backward (\(\beta\)) probabilities, which are crucial for evaluating sequence likelihoods and performing gradient-based optimization during model training. By leveraging the parallel processing capabilities of GPUs, the algorithm distributes computations across numerous threads and blocks, significantly reducing runtime compared to traditional CPU-based implementations.

\subsection{Dynamic Programming Framework: Forward Probability (\(\alpha\)) and Backward Probability (\(\beta\))}

The forward algorithm computes the forward probability matrix \(\alpha\), where each element \(\alpha_{i,u}\) represents the probability of reaching state \(u\) at position \(i\) in the target sequence. The computation follows the recursive relation:

\[
\alpha_{i,u} = \sum_{v < u} \left( \alpha_{i-1,v} \times \mathbf{E}_{v,u} \times \mathbf{P}_{u}(y_i) \right),
\]

where \(v\) and \(u\) index the states, \(\mathbf{E}_{v,u}\) is the transition score from state \(v\) to state \(u\), \(\mathbf{P}_{u}(y_i)\) is the emission probability of state \(u\) emitting the \(i\)-th token \(y_i\).

\paragraph{CUDA Parallelization Strategy for the Forward Algorithm.} Each GPU thread is assigned to compute \(\alpha_{i,u}\) for specific states and positions, allowing multiple \(\alpha_{i,u}\) values to be calculated concurrently. Warp-level optimizations, are employed to efficiently perform reductions when summing contributions from multiple states. Additionally, shared memory is utilized to store intermediate results, and synchronization queues coordinate computations across different segments of the sequence. This parallel approach enables the simultaneous computation of numerous \(\alpha_{i,u}\) values, thereby accelerating the forward pass across all sequence positions and states.

The backward algorithm calculates the backward probability matrix \(\beta\), where each element \(\beta_{i,u}\) signifies the probability of transitioning from state \(u\) at position \(i\) to the end of the sequence. The recursive relation for \(\beta\) is defined as:

\[
\beta_{i,u} = \sum_{v > u} \left( \beta_{i+1,v} \times \mathbf{E}_{u,v} \times \mathbf{P}_{v}(y_{i+1}) \right),
\]

where \(v\) and \(u\) are state indices, \(\mathbf{E}_{u,v}\) is the transition score from state \(u\) to state \(v\), \(\mathbf{P}_{v}(y_{i+1})\) is the emission probability of state \(v\) emitting the \((i+1)\)-th token \(y_{i+1}\).

\paragraph{CUDA Parallelization Strategy for the Backward Algorithm.} Similar to the forward pass, GPU threads are allocated to compute \(\beta_{i,u}\) for specific states and positions. The backward pass processes the sequence in reverse order, ensuring that \(\beta_{i+1,v}\) values are computed before \(\beta_{i,u}\) values that depend on them. Warp-level primitives facilitate efficient summation of probabilities from future states, while shared memory and synchronization mechanisms maintain data consistency across different sequence segments. This parallelization allows for the rapid aggregation of probabilities from multiple future states, effectively computing the entire \(\beta\) matrix.

\subsection{Gradient Calculations}

Gradient computation is essential for optimizing model parameters during training. The algorithm calculates gradients with respect to both emission probabilities \(\mathbf{P}_{u}(y_i)\) and transition scores \(\mathbf{E}_{v,u}\). In the following formulas, \(M\) denotes the final position in the target sequence (i.e., the last index along the target length dimension).

  \[
  \nabla \mathbf{P}_{u}(y_i) = \frac{\alpha_{i,u} \times \beta_{i,u}}{\sum_{u} \alpha_{M,u}}
  \]

where \(\sum_{u} \alpha_{M,u}\) represents the likelihood of the entire sequence, computed by summing the forward probabilities \(\alpha_{M,u'}\) over all possible final states \(u'\) at the final position \(M\). This gradient indicates how changes in the emission probability for state \(u\) at position \(i\) affect the log-likelihood of the sequence.

The transition gradient is given by:

  \[
  \nabla \mathbf{E}_{v,u} = \frac{\alpha_{i-1,v} \times \beta_{i,u}}{\sum_{u} \alpha_{M,u}}
  \]

This gradient measures the effect of changes in the transition score between states \(v\) and \(u\) on the overall log-likelihood of the sequence.

\paragraph{CUDA Parallelization Strategy.} For emission gradients, each thread independently computes \(\nabla \mathbf{P}_{u}(y_i)\) for assigned states and positions, leveraging parallel execution to handle multiple computations simultaneously. For transition gradients, threads collaboratively compute \(\nabla \mathbf{E}_{v,u}\) by aggregating contributions from various sequence positions using parallel reduction techniques. This parallel approach ensures efficient backpropagation through the dynamic programming steps, facilitating rapid parameter updates during training.

\subsection{Time Complexity Analysis}

\paragraph{Serial (Naïve) Time Complexity.} In a sequential implementation, computing the forward and backward probabilities involves iterating over all batches, sequence positions, states, and transitions. The total number of operations scales as:

\[
\mathcal{O}\left(\text{bsz} \times \text{tarlen} \times \text{prelen} \times \text{translen}\right),
\]

where \(\text{bsz}\) is batch size, \(\text{tarlen}\) is target sequence length, \(\text{prelen}\) is the number of states, \(\text{translen}\) is the maximum number of transitions per state (usually equals to \(\text{prelen}\)).

\paragraph{CUDA Parallel Time Complexity.} Under ideal parallel conditions, assuming the GPU can handle an effectively infinite number of parallel threads, which means that each sequence position can be processed in parallel during the forward and backward passes, reducing the time complexity to $\mathcal{O}\left(\text{tarlen}\right)$, since each position \(i\) can be computed independently once the necessary dependencies are met. The gradient computations similarly benefit from parallel execution, allowing gradients for different states and transitions to be calculated concurrently.

While the theoretical parallel time complexity suggests \(\mathcal{O}(\text{tarlen})\), actual performance is influenced by GPU resource constraints, such as the number of available threads, memory bandwidth, and synchronization overhead. In practice, the wall-clock time is significantly reduced compared to the serial case, though it may scale slightly worse than \(\mathcal{O}(\text{tarlen})\) due to these hardware and implementation factors.

\subsection{Space Complexity Analysis}

The algorithm's space requirements are primarily determined by the storage of the \(\alpha\) and \(\beta\) matrices, along with auxiliary tensors used for transition scores. The overall space complexity is:

\[
\mathcal{O}\left(\text{bsz} \times \text{prelen} \times (\text{tarlen} + \text{translen})\right),
\]

where \(\text{bsz}\) is batch size, \(\text{tarlen}\) is target sequence length, \(\text{prelen}\) is the number of states, \(\text{translen}\) is the maximum number of transitions per state (usually equals to \(\text{prelen}\)).

This accounts for the forward and backward probability matrices and the transition scores necessary for computations.

\paragraph{Real-World Analysis of Algorithm Space Usage.} In practical scenarios, the space requirements of the CUDA-accelerated dynamic programming algorithm remain manageable, even for large-scale tasks such as masked sequence modeling for text. For instance, consider a typical setup with a batch size of $64$ per GPU, a maximum sequence length of $512$ tokens, and a masking ratio of $0.15$, resulting in approximately $77$ \mt{} tokens per sequence. If each \mt{} token can be expanded to up to $4$ hidden states ($k=4$), the prelen (number of states) becomes approximately \(512 \times 0.15 \times 4 \approx 308\). Similarly, tarlen (the target sequence length) is approximately $77$, while translen (the number of transitions per state) is set to $308$, matching prelen. Under these conditions, tensors such as \(\alpha\) and \(\beta\), each of size \([\text{bsz}, \text{tarlen}, \text{prelen}]\), would require around \(64 \times 77 \times 308 \approx 1.52 \times 10^6\) entries, equivalent to $5.8$ MB per tensor for 32-bit floats. The transition links tensor, of size \([\text{bsz}, \text{prelen}, \text{translen}]\), would require approximately \(64 \times 308 \times 308 \approx 6.07 \times 10^6\) entries, or about $23.2$ MB of memory for 32-bit floats. Altogether, these tensors occupy approximately $34.8$ MB of memory. Even with these allocations, the memory usage is well within the $80$ GB available on an NVIDIA A100 GPU, leaving ample room for model parameters, activations, and framework overheads. \textbf{This demonstrates that the algorithm's \(\displaystyle \mathcal{O}(\text{bsz} \times \text{prelen} \times (\text{tarlen} + \text{translen}))\) space complexity poses no significant constraint in real-world training setups}.

\paragraph{States alignment algorithm does not affect the scalability of the model.} The analysis of the time and space complexity of the states alignment algorithm reveals that its complexity is independent of the model size (e.g., the number of layers or the embedding dimension) and depends only on the input sequence length. Therefore, with the training data unchanged, states alignment algorithm does not introduce additional training overhead as the model size increases. 


\section{Hyper-Parameter Configuration for Molecular Pre-training}
\label{sec::pretrain_configuration}

We implement \method{} using $9$ stacked Transformer layers, each with $12$ attention heads. The model dimension and feedforward dimension of each Transformer layer are $768$ and $2{,}048$, respectively. The total number of \method{}'s parameters achieves $50.5$M. We use Adam \citep{kingma2014adam} optimizer and polynomial learning rate scheduler to train \method{}, and we set the learning rate as 5e-4 and warmup stesp as $10$K.
The total training steps are $120$K and each batch has $64$k tokens at maximum. We implement the \method{} model using the Fairseq library \footnote{\url{https://fairseq.readthedocs.io/en/latest/}} and train \method{} on two RTX3090 GPUs for about 24 hours.

For more pre-training hyper-parameters, please refer to Table \ref{table_hyper}.

\begin{table}[ht]
\footnotesize
\color{black}{
\caption{\method{} hyper-parameters for molecular pre-training.}
\label{table_hyper}

\begin{center}
\begin{tabular}{ccc}
\toprule
\multicolumn{1}{c}{\multirow{1}{*}{Hyper-parameters}} &\multicolumn{1}{c}{\multirow{1}{*}{Value}}\\
\midrule
Learning rate  & 5e-4 \\
LR scheduler  & polynomial\_decay \\
Num of expanded states & $2$ \\
Warmup updates & $10$K \\
Max updates & $120$K \\
Max tokens & $64$K \\
FFN dropout & $0.1$ \\
Attention dropout & $0.1$ \\
Activation dropout & $0$ \\
Num of layers & $9$ \\
Num of attention heads & $12$ \\
Encoder embedding dim & $768$ \\
Encoder FFN dim & $2,048$ \\
Adam ($\beta_1, \beta_2$) & $(0.9,0.98)$ \\
Fragments Drop ratio &  $0.15$ \\
Vocabulary size & $369$ \\
Activation function & GELU \\
Weight Decay & $0.0$ \\
Clip Norm & $1.0$ \\
\bottomrule
\end{tabular}
\end{center}
}
\end{table}

\section{Hyper-Parameter Configuration for Molecular Fine-tuning}
\label{sec::ft_configuration}

In different downstream task, we use different hyper-parameters. 
We run each task three times using three different random seeds, and take the average performance of these three runs as the final result. For detailed fine-tuning hyper-parameters, please refer to Table \ref{table_hyper_ft}.
\begin{table}[ht]
\footnotesize
\color{black}{
\caption{\method{} hyper-parameters for molecular fine-tuning.}
\label{table_hyper_ft}

\begin{center}
\begin{tabular}{ccccccc}
\toprule
\multicolumn{1}{c}{\multirow{1}{*}{Tasks}} &\multicolumn{1}{c}{\multirow{1}{*}{Epochs}} &\multicolumn{1}{c}{\multirow{1}{*}{Batch size}} &\multicolumn{1}{c}{\multirow{1}{*}{Learning rate}} &\multicolumn{1}{c}{\multirow{1}{*}{Warmup Ratio}} &\multicolumn{1}{c}{\multirow{1}{*}{Dropout}} &\multicolumn{1}{c}{\multirow{1}{*}{Pooler-dropout}} \\
\midrule

BACE & 60 & 64 & 1e-4 & 0.06 & 0.1 & 0.2 \\
BBBP & 40 & 128 & 4e-4 & 0.06 & 0.1 & 0.1 \\
TOX21 & 80 & 128 & 1e-4 & 0.06 & 0.1 & 0.1 \\
SIDER & 100 & 32 & 5e-4 & 0.4 & 0.1 & 0 \\
MUV & 40 & 128 & 2e-5 & 0.2 & 0.1 & 0.1 \\
ClinTox & 100 & 256 & 5e-5 & 0.1 & 0.1 & 0.5 \\
ToxCast & 80 & 64 & 1e-4 & 0.06 & 0.1 & 0.1 \\
\bottomrule
\end{tabular}
\end{center}
}
\end{table}


\section{Details of Molecular Fine-tuning Datasets and Baselines}
\label{sec::ft_datasets}
\paragraph{Datasets.} We perform a comprehensive set of experiments on the MoleculeNet\citep{wu2018moleculenet} benchmark, focusing on the molecular property prediction task. MoleculeNet has emerged as one of the most widely recognized and utilized benchmarks in the field of molecular property prediction, providing a standardized platform for evaluating machine learning models' performances on evaluating molecular properties. Its datasets encompass a broad range of molecular tasks, and address diverse and practical scientific problems such as drug discovery, toxicity prediction and so on.

In this section, we provide a detailed summary of the statistics and fundamental characteristics of the MoleculeNet datasets we use in Table \ref{fine_tune_dataset}. This table offers information about the dataset sizes, task types, and compositions, providing readers with essential background information to better understand the experimental setup and subsequent analysis.

\paragraph{Baselines.}
 We evaluate our approach against various supervised learning and pre-training baselines, including both SMILES-based and 3D molecular pre-trained models. The supervised methods include D-MPNN \citep{yang2019analyzing} and AttentiveFP \citep{xiong2019pushing}, both of which are based on graph neural networks (GNNs). For 2D and 3D molecular pre-training, we consider baseline methods: N-gram \citep{liu2019n}, GROVER \citep{rong2020self}, GraphMVP \citep{liu2021pre}, Mole-BERT \citep{xiamole}, and 3D InfoMax \citep{stark20223d}. For a fair comparison, we train a SMILES model based on MLM pre-training, referred to as SMILES-BERT, using the same training data, model architecture, and training hyperparameters as \method{}.

\begin{table*}[ht]
\caption{Summary information of the MoleculeNet benchmark datasets.\label{tab1}}
\tabcolsep=0pt
\begin{tabular*}{\textwidth}{@{\extracolsep{\fill}}ccccp{0.25\textwidth}@{\extracolsep{\fill}}}
\toprule
Dataset & Tasks & Task type & Molecules (train/valid/test) &  Describe \\
\midrule
BACE &  1 & Classification & 1,210/151/151 & Binding results of human BACE-1 inhibitors \\
BBBP &  1 & Classification  & 1,631/204/204 & Blood-brain barrier penetration \\
ClinTox &  2 & Multi-label classification  & 1,182/148/148 & Clinical trial toxicity and FDA approval status \\
Tox21 &  12 & Multi-label classification  & 6,264/783/783 & Qualitative toxicity measurements \\
ToxCast &  617 & Multi-label classification & 6,860/858/858 & Toxicology data based on in vitro screening \\
SIDER &  27 & Multi-label classification & 1,141/143/143 & Adverse drug reactions to the 27 systemic organs \\
MUV &  17 & Multi-label classification & 74,469/9,309/9,309 & A subset of PubChem BioAssay \\
\bottomrule
\end{tabular*}
\label{fine_tune_dataset}
\end{table*}

\clearpage
\section{Details of the GLUE Benchmark}
\label{app:glue}

Below are detailed descriptions of all the GLUE benchmark tasks, which collectively evaluate various aspects of natural language understanding such as entailment, paraphrase detection, sentiment analysis, and grammaticality judgment:

\paragraph{MNLI (Multi-Genre Natural Language Inference).}The MNLI dataset~\citep{MNLI} consists of $393$K training examples gathered through crowdsourcing from various genres. The task requires predicting whether a premise sentence entails, contradicts, or is neutral with respect to a given hypothesis sentence.

\paragraph{QQP (Quora Question Pairs).}QQP~\citep{QQP} includes $364$K training examples sourced from the Quora question-answering platform. The objective is to determine if two provided questions are semantically equivalent.

\paragraph{QNLI (Question Natural Language Inference).}Derived from the Stanford Question Answering Dataset (SQuAD)~\citep{Rajpurkar2018KnowWY}, QNLI comprises $108$K training examples. The task involves predicting whether a sentence contains the answer to a specific question.

\paragraph{SST-2 (Stanford Sentiment Treebank).}SST-2~\citep{SST-2} contains $67$K training examples based on movie reviews with human-annotated sentiments. The goal is to classify each sentence as expressing either positive or negative sentiment.

\paragraph{CoLA (Corpus of Linguistic Acceptability).}The CoLA dataset~\citep{COLA} includes $8.$K training examples extracted from books and journal articles focused on linguistic theory. The task is to assess whether a given sentence is linguistically acceptable.

\paragraph{RTE (Recognizing Textual Entailment).}RTE~\citep{RTE-5,RTE-1,RTE-2,RTE-3} encompasses $2.5$K training examples derived from textual entailment challenges. The objective is to determine if a premise sentence entails a provided hypothesis sentence.

\paragraph{MRPC (Microsoft Research Paraphrase Corpus).}MRPC~\citep{MRPC} consists of $3.7$K training examples collected from various news sources. The task is to predict whether two given sentences are semantically equivalent.

\paragraph{STS-B (Semantic Textual Similarity Benchmark).}STS-B~\citep{STS-B} includes $5.8$K training examples sourced from multiple origins, annotated by humans for sentence pair semantic similarity. The task requires predicting the degree of semantic similarity between two sentences on a scale from $1$ to $5$.

We use Spearman correlation for STS, Matthews correlation for CoLA, and accuracy for MNLI, QNLI, RTE and SST-2 as the metrics on GLUE.

\clearpage

\section{Hyper-Parameter Configuration for Textual Pre-training}
\label{sec::text_pretrain_configuration}

We implement \method{} in two configurations: a base model and a large model. The base \method{} consists of 12 stacked Transformer layers, each with 12 attention heads. The model dimension and feedforward dimension of each Transformer layer are $768$ and $3{,}072$, respectively, resulting in a total of $128$M parameters. The large \method{} model uses $24$ Transformer layers with $16$ attention heads per layer. The model dimension and feedforward dimension are increased to $1,024$ and $4,096$, respectively, with a total parameter count of $361$M. We use Adam \citep{kingma2014adam} optimizer and polynomial learning rate scheduler to train \method{}, and we set the learning rate as 5e-4 and warmup steps as $10$K.
The total training steps are $125$K and each batch has $2048$ samples at maximum. We also implement the \method{} model using the Fairseq library. For more pre-training hyper-parameters, please refer to Table \ref{text_table_hyper}.

\begin{table}[ht]
\footnotesize
\color{black}{
\caption{\method{} hyper-parameters for textual pre-training.}
\label{text_table_hyper}

\begin{center}
\begin{tabular}{ccc}
\toprule
\multicolumn{1}{c}{\multirow{1}{*}{Hyper-parameters}} &\multicolumn{1}{c}{\multirow{1}{*}{Value}}\\
\midrule
Learning rate  & 5e-4 \\
LR scheduler  & polynomial\_decay \\
Num of expanded states & $4$ \\
Warmup updates & $10$K \\
Max updates & $125$K \\
Batch size & $2,048$ \\
FFN dropout & $0.1$ \\
Attention dropout & $0.1$ \\
Activation dropout & $0$ \\
Num of layers & \textit{base}: $12$, \textit{large}: $24$  \\
Num of attention heads & \textit{base}: $12$, \textit{large}: $16$ \\
Encoder embedding dim & \textit{base}: $768$, \textit{large}: $1,024$ \\
Encoder FFN dim & \textit{base}: $3,072$, \textit{large}: $4,096$ \\
Adam ($\beta_1, \beta_2$) & $(0.9,0.98)$ \\
Mask ratio &  $0.15$ \\
Activation function & GELU \\
Weight Decay & $0.01$ \\
Clip Norm & $0.0$ \\
\bottomrule
\end{tabular}
\end{center}
}
\vspace{-0.5cm} 
\end{table}

\section{Hyper-Parameter Configuration for Textual Fine-tuning}
\label{app:hyper}

We apply grid search for both the GLUE and SQuAD 2.0 datasets, and the grid search hyperparameters are shown in Table~\ref{tab:glue_hyper}. 

\begin{table*}[h]
    \centering
    \caption{Grid search hyperparameters for the GLUE and SQuAD 2.0 tasks.}
    \begin{tabular}{lc}
        \toprule
        \textbf{Hyperparameter} & MNLI, QNLI, QQP, SST-2  \\
        \midrule
        Peak learning rate & \{1e-5, 2e-5, 3e-5, 4e-5\} \\
        Batch size &32  \\
        Max epochs & \{2, 3, 5\} \\
        Warm-Up Proportion & 6\% \\
        \midrule
        & RTE, MRPC, CoLA, STS-B \\
        \midrule
        Peak learning rate &  \{2e-5, 3e-5, 4e-5, 5e-5\} \\
        Batch size & \{16, 32\} \\
        Max epochs & \{2, 3, 5, 10\}\\
        Warm-Up Proportion & \{6\%, 10\%\}\\
        \midrule
        & SQuAD 2.0 \\
        \midrule
        Peak learning rate &  \{2e-5, 3e-5, 4e-5, 5e-5\} \\
        Batch size & \{16, 32\} \\
        Max epochs & \{2, 3\}\\
        Warm-Up Proportion & \{6\%, 10\%\}\\
        \bottomrule
    \end{tabular}
    
    \label{tab:glue_hyper}
\end{table*}

\clearpage

\section{Detailed Case Study}
\label{app:case_study}

\begin{figure*}[h]
\centering
\includegraphics[width=1.0\linewidth]{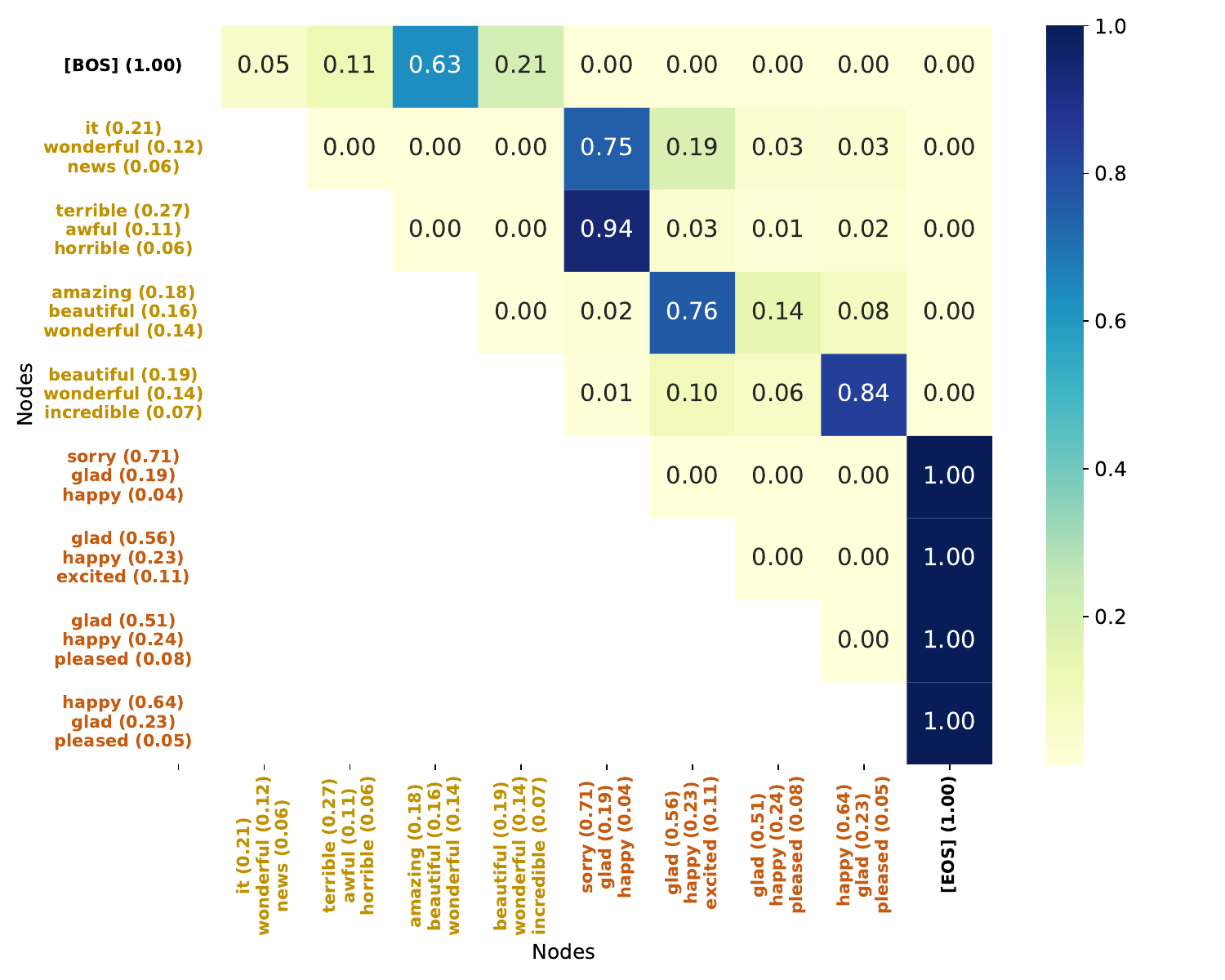}
\caption{We visualize the model's predictions when the input is ``\textit{This is \mt{}, and I’m very \mt{} to see this}." ($k=4$). The \textcolor[HTML]{BF9000}{\textbf{yellow nodes}} represent the expanded states corresponding to the first \mt{} token, while the \textcolor[HTML]{C55A11}{\textbf{brown nodes}} represent the expanded states corresponding to the second \mt{} token. The x-axis and y-axis show the top-3 predicted word for each graph node.}
\label{app:case_heatmap}
\end{figure*}

\begin{figure*}[h]
\centering
\includegraphics[width=1.0\linewidth]{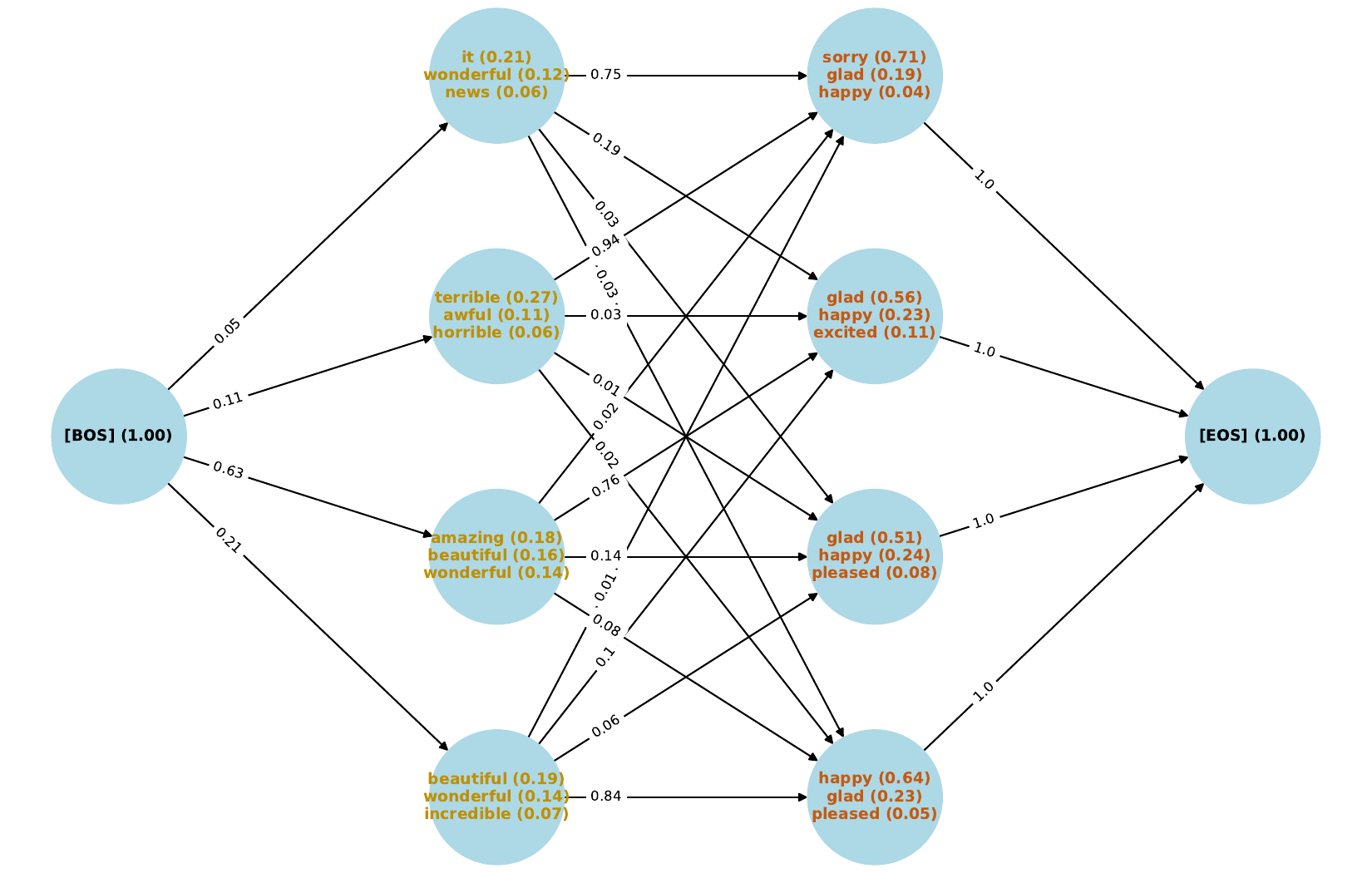}
\caption{The visualization of the DAG from \method{}. It shows that the edge weight between different nodes is directly related to the semantic dependency between those nodes.}
\label{app:case_dag}
\end{figure*}

We further provide a visualization of the model's predictions when the input is \textit{This is \mt{}, and I’m very \mt{} to see this.}, as shown in Figure \ref{app:case_heatmap}. Specifically, since the input contains two \mt{} tokens and \(k=4\), there are 8 nodes in the graph. For implementation convenience, we introduce two special nodes, \texttt{[BOS]} and \texttt{[EOS]}, which correspond to the starting and ending nodes of the graph. We require that all paths in the graph must start from the \texttt{[BOS]} node and terminate at the \texttt{[EOS]} node, so when searching for a valid path in the graph, we can directly start from the \texttt{[BOS]} node without enumerating all possible starting nodes. In this graph, besides the heatmap representing the transition probability (i.e., the edge weight between two nodes), we also display the top 3 most probable words decoded from each graph node. From these results, we observe the following characteristics of \method{}:

\begin{itemize}
    \item \textbf{\method{} successfully represents the different choices of each \mt{} token using distinct expanded states.} Since \(k=4\), the model predicts four different choices for both \mt{} tokens in the input. These four choices typically represent different aspects of the semantics. For example, the first \mt{} token's choices include three categories: a noun (e.g., it, news), negative emotion words (e.g., terrible, awful), and positive emotion words (e.g., amazing, beautiful, wonderful), each represented by a different expanded state. This effectively avoids semantic ambiguity in the context, i.e., the semantic multimodality. Similarly, the second \mt{} token's choices include positive emotions (e.g., glad, happy) and negative emotions (e.g., sorry), and these different types of semantic information are also represented by distinct expanded states.
    \item \textbf{\method{} successfully captures the semantic relationships between different \mt{} tokens.} The transition matrix displayed in Figure \ref{app:case_heatmap} shows an important phenomenon: the degree of dependency between different states in \method{} (i.e., the edge weight between two nodes) is strongly correlated with the semantic similarity between them. For instance, when the first \mt{} token is "terrible," the second \mt{} token has a higher probability of choosing "sorry." Conversely, when the first \mt{} token is a more positive emotion word, the second \mt{} token is more likely to choose "glad," which also represents positive emotion. This demonstrates that \method{} can effectively learn the semantic dependencies between \mt{} tokens. We further visualize this as a directed acyclic graph (DAG) in Figure \ref{app:case_dag}, where the edge weight between nodes is directly related to the semantic dependency between them, further validating the effectiveness of \method{}.
    \item \textbf{\method{} typically has lower uncertainty for \mt{} tokens that appear later}. As shown in Figure \ref{app:case_heatmap}, the word probability distributions of the four expanded states corresponding to the \textcolor[HTML]{BF9000}{\textbf{first \mt{} token}} are often more evenly distributed, meaning that the differences of word probabilities across different words in each state are not very large. However, for the four expanded states corresponding to the \textcolor[HTML]{C55A11}{\textbf{second \mt{} token}}, the model usually predicts one word with a probability significantly higher than the others. This phenomenon occurs due to the directionality of the DAG. Specifically, in \method{}'s DAG, transitions only occur from earlier states to later states, and the choices of earlier states have a greater impact on the final result, which means that the earlier states tend to have higher uncertainty.
\end{itemize}

\clearpage

\section{More Analytical Experiments of \method{}}
\label{app:more_analysis_exmlm}

\subsection{Entropy Analysis on \method{}}

\begin{figure*}[h]
\centering
\includegraphics[width=0.7\linewidth]{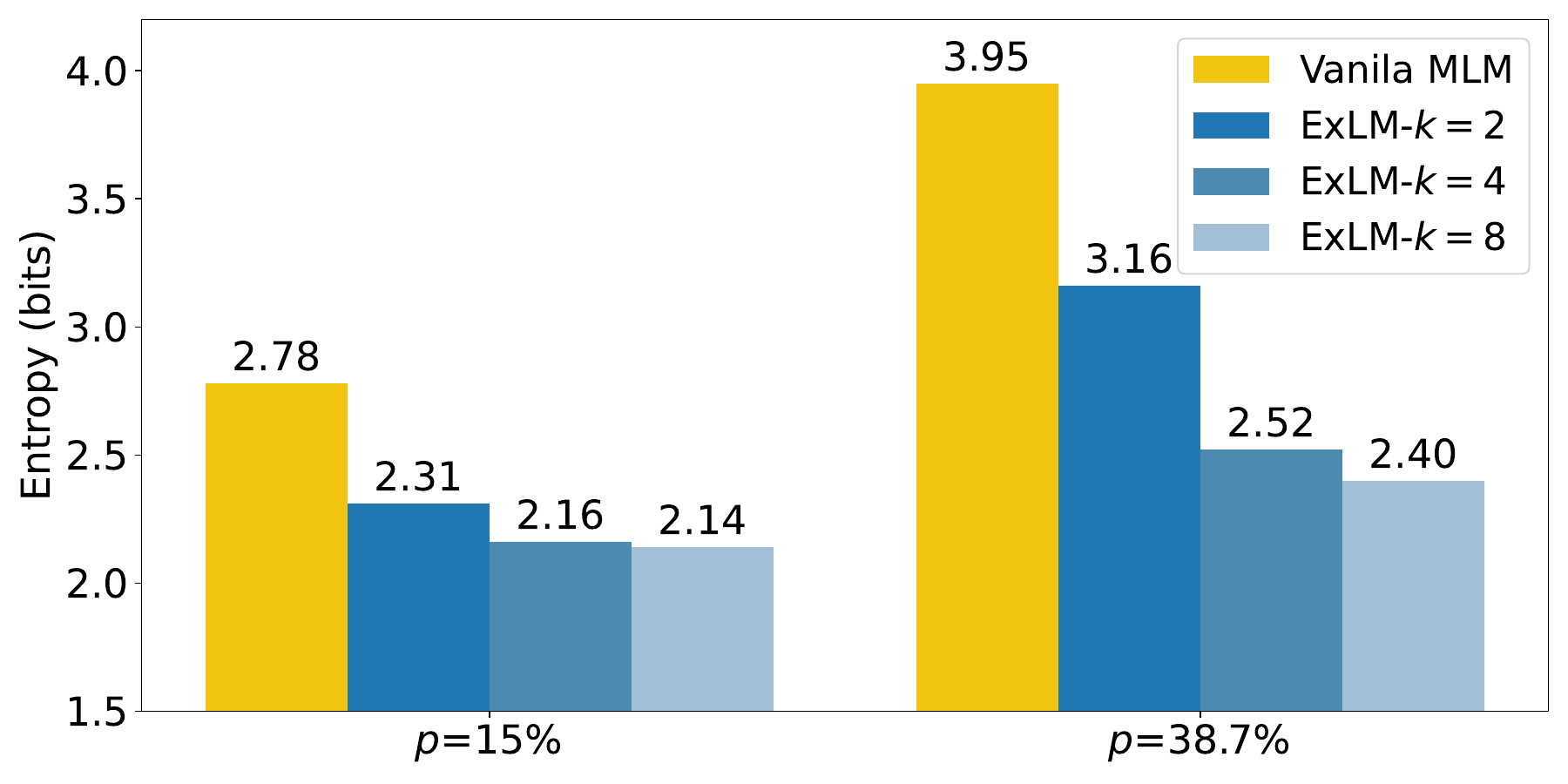}
\caption{Entropy analysis of \method{} with different mask ratios $p$ and numbers of expanded states \(k\). \method{} demonstrates significantly lower uncertainty than MLM, and a larger \(k\) further reduces the \method{}'s uncertainty, enhancing its better modeling capability.}
\label{app:ent_exmlm}
\end{figure*}

We conduct an entropy analysis to compare the severity of the semantic multimodality under different mask ratios \(p\) in the input context for \method{} and MLM. The results are shown in Figure \ref{app:ent_exmlm}. The results reveal that \method{} has significantly lower entropy in its prediction distribution than MLM, indicating lower uncertainty in dealing with ambiguous contexts. This further proves that \method{} significantly mitigates the negative impact of semantic multimodality and can still produce relatively confident predictions even when contextual semantic corruption is severe. Additionally, as \(k\) increases, the uncertainty of \method{} decreases, suggesting that increasing \(k\) expands the model's semantic space, helping to alleviate the negative effects of the semantic multimodality. Moreover, under higher mask ratios, \method{} shows a more noticeable reduction in uncertainty compared to MLM, explaining why \method{} still maintains relatively strong performance under high mask ratios.

\subsection{Training Efficiency of \method{}}

We compare the training efficiency differences between MLM and \method{} ($k=4$) with the same pre-training configuration. Both models are trained on two Tesla A100 80G GPUs under the hyperparameters from Table \ref{analyse_table_hyper}, and their training cost statistics are shown in Table \ref{gpu_time}. From this table, we can know that the training cost of \method{} ($k=4$) is approximately 1.9 times that of the MLM with the same pre-training configuration. To compare the performance differences between MLM and \method{} ($k=4$) under equal training costs, we increase the training steps of MLM from $50{,}000$ to $100{,}000$, and refer to this model as Vanilla MLM++. The performance of Vanilla MLM++ is shown in Table \ref{table::ablations}, where we can see that although increasing the training steps significantly improves MLM performance, it still remains lower than that of \method{}. This further demonstrates that \method{} achieves stronger performance under the same training cost.

\begin{table}[ht]
\footnotesize
\color{black}{
\caption{Training time comparison (Tesla A100 80G GPUs) between \method{} ($k=4$) and MLM.}
\label{gpu_time}

\begin{center}
\begin{tabular}{cccc}
\toprule
\multicolumn{1}{c}{\multirow{1}{*}{ }} &\multicolumn{1}{c}{\multirow{1}{*}{MLM}}&\multicolumn{1}{c}{\multirow{1}{*}{\method{} ($k=4$)}}\\
\midrule
GPU Time (Hours)  & 54.7 & 104.2 \\
\bottomrule
\end{tabular}
\end{center}
}
\end{table}

\subsection{Performance of Sparse \method{}}

We also implement a version of \method{} using a Sparse DAG (Sparse \method{}) and evaluate its performance on various downstream tasks, with the results shown in Table \ref{table::ablations_sparse}. Specifically, the key difference between the Sparse DAG and the original DAG in \method{} is that, in the Sparse DAG, we do not allow transitions between expanded states that belong to the same \mt{} token. This significantly reduces the number of edges in the DAG, making the graph more sparse. 

However, as shown in Table \ref{table::ablations_sparse}, \method{} with the Sparse DAG experiences some performance degradation. While it still outperforms the version of \method{} without a transition matrix (\method{} w/o Transitions), it falls behind the original \method{}. The reason for this decline is that, in the Sparse DAG, only one expanded state can be selected for each \mt{} token, which prevents the model from capturing longer but possible outputs. For example, in the sentence ``\textit{This is \mt{}, and I’m very \mt{} to see this}," both ``\textit{so good}" and ``\textit{happy}" are reasonable interpretations for the first and second \mt{} tokens, respectively. However, the Sparse DAG cannot model this ambiguity, as it does not allow multiple expanded states being chosen for a single \mt{} token. This limitation significantly restricts the model's ability to capture rich semantic variations, leading to a performance drop.

\begin{table}[h]
\centering
\caption{
\textbf{Performance of Sparse \method{}.} Using a sparser DAG in \method{} leads to a performance decline. 
}
\begin{tabular}{c|cccc|c}
  \toprule
  Method  & MNLI $\uparrow$ & QNLI$\uparrow$ & QQP$\uparrow$ & RTE$\uparrow$ & Avg  $\uparrow$\\
  \midrule
  
 \method{} w/o Transitions  & 83.8 & 90.9 & 91.1 & 55.6 & 80.4 \\
 \method{} w/ Sparse DAG & 84.4 & 91.2 & 91.3 & 56.9 & 81.0 \\
  \midrule 
  Vanilla MLM  & 83.6 & 90.0 & 90.3 & 54.7 & 79.6\\
  \midrule 
  \method{}  & \textbf{85.1} & \textbf{91.4} & \textbf{91.3} & \textbf{57.6} & \textbf{81.4} \\
  \bottomrule
\end{tabular}
\label{table::ablations_sparse}
\end{table}


\end{document}